\documentclass[10pt,journal,compsoc]{IEEEtran}

\ifCLASSOPTIONcompsoc
  \usepackage[nocompress]{cite}
\else
  \usepackage{cite}
\fi

\ifCLASSINFOpdf

\else

\fi

\usepackage{times}
\usepackage{epsfig}
\usepackage{graphicx}
\usepackage{amsmath}
\usepackage{amssymb}

\usepackage{refstyle}
\usepackage{bm} 
\usepackage{subcaption}
\usepackage{appendix}
\usepackage[font={small}]{caption}
\usepackage[abs]{overpic}
\usepackage{bbding}
\usepackage{mathrsfs}
\usepackage[T1]{fontenc}
\usepackage[utf8]{inputenc}

\usepackage{xcolor}
\usepackage{adjustbox}
\usepackage{ragged2e}  
\usepackage{multirow}  
\usepackage{gensymb}  
\usepackage{soul,color}  

\usepackage[breaklinks=true,bookmarks=false]{hyperref}

\def\ie{i.e., }

\newcommand{\words}[1]{{\color{black}#1}}
\newcommand{\sentences}[1]{{\color{black}#1}}

\begin{document}






%
\title{SurfaceNet+: An End-to-end 3D Neural Network \\ for Very Sparse Multi-view Stereopsis}
%
%
%
%

\author{Mengqi~Ji$^*$, 
        Jinzhi~Zhang$^*$, 
        Qionghai~Dai, 
        Lu~Fang$^{\S}$
        \par 
        Tsinghua University \thanks{
        \newline
        $*:$ Equal Contribution
        \newline
        $\S:$ Corresponding Author (\href{mailto:fanglu@sz.tsinghua.edu.cn}{fanglu@sz.tsinghua.edu.cn},
        \url{http://luvision.net})
        \newline
        \newline
        This work is supported in part by Natural Science Foundation of China (NSFC) under contract No. 61722209 and 6181001011, in part by Shenzhen Science and Technology Research and Development Funds (JCYJ20180507183706645).
        
        }
        }

\markboth{}%
{Shell \MakeLowercase{\textit{et al.}}: Bare Demo of IEEEtran.cls for Computer Society Journals}
%



\IEEEtitleabstractindextext{%
\justify  
\begin{abstract}

Multi-view stereopsis (MVS) tries to recover the 3D model from 2D images. As the observations become sparser, the significant 3D information loss makes the MVS problem more challenging. Instead of only focusing on densely sampled conditions, we investigate sparse-MVS with large baseline angles since \sentences{the sparser sensation is more practical and more cost-efficient}. 
By investigating various observation sparsities, we show that the classical depth-fusion pipeline becomes powerless for the case with \words{a} larger baseline angle that worsens the photo-consistency check. As another line of \words{the} solution, we present SurfaceNet+, a volumetric method to handle the `incompleteness' and \words{the} `inaccuracy' problems induced by \words{a} very sparse MVS setup. Specifically, the former problem is handled by a novel volume-wise view selection approach. It owns superiority in selecting valid views while discarding invalid occluded views by considering the geometric prior. Furthermore, the latter problem is handled via a multi-scale strategy that consequently refines the recovered geometry around the region with \words{the} repeating pattern. The experiments demonstrate the tremendous performance gap between SurfaceNet+ and state-of-the-art methods in terms of precision and recall. Under the extreme sparse-MVS settings in two datasets, where existing methods can only return very few points, SurfaceNet+ still works as well as in the dense MVS setting. 

\end{abstract}


}

\maketitle

\IEEEdisplaynontitleabstractindextext

%
\IEEEpeerreviewmaketitle

\section{Introduction}\label{sec:introduction}
\IEEEPARstart{M}{ulti-view} stereopsis (MVS) aims to recover a dense 3D model from a set of 2D images with known camera parameters. As the observations become sparser, the more 3D information of the imaged scene get lost during the sensing procedure, making the following perception procedure, for example, an MVS task, more challenging. \sentences{Dense multi-view sensation has attracted tremendous attention in light field imaging and rendering. Its advantages, such as being robust to occlusion \mbox{\cite{yucer2016depth}\cite{wu2018light}} and reducing image noise \mbox{\cite{bishop2009light}\cite{chung2019computational}}, have been well studied. Unfortunately, it is impractical to densely sample a scene for high-resolution 3D reconstruction, especially for the large-scale scenes. In contrast, the sparser sensation with a wide baseline is more practical and more cost-efficient; however, it aggravates the difficulty of MVS problem since the larger baseline angles lead to tough dense-correspondence matching.}

\begin{figure}[t]
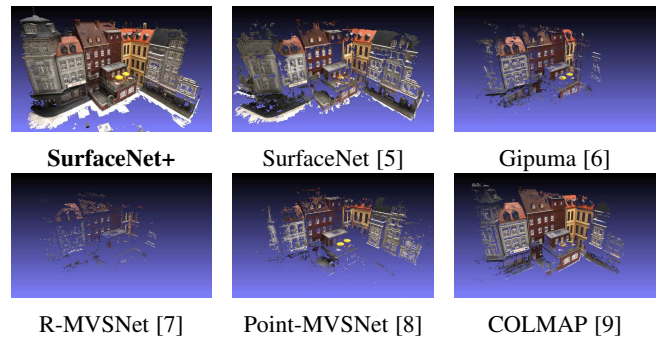

    \centering
    \newcommand{\colw}{0.30}
    \newcommand{\figw}{1.0} 

    \begin{subfigure}[t]{\colw\linewidth}
        \begin{minipage}{\textwidth}
            \includegraphics[width=1\linewidth,  keepaspectratio=true,trim={0cm 0cm 0cm 0cm},clip]{{{figures/results/compare/DTU_sparse/model_23/surface+00_L03}}}
           
        \end{minipage}
        \captionsetup{labelformat=empty}
         \caption{\textbf{SurfaceNet+}}
    \end{subfigure}
    ~ 
    \begin{subfigure}[t]{\colw\linewidth}
        \begin{minipage}{\textwidth}
            \includegraphics[width=1\linewidth,  keepaspectratio=true,trim={0cm 0cm 0cm 0cm},clip]{{{figures/results/compare/DTU_sparse/model_23/surface0500}}}
           
        \end{minipage}
        \captionsetup{labelformat=empty}
        \caption{SurfaceNet\cite{ji2017surfacenet}}
    \end{subfigure}
    ~ 
    \begin{subfigure}[t]{\colw\linewidth}
        \begin{minipage}{\textwidth}
            \includegraphics[width=1\linewidth,  keepaspectratio=true,trim={0cm 0cm 0cm 0cm},clip]{{{figures/results/compare/DTU_sparse/model_23/gipuma00_L03}}}
           
        \end{minipage}
        \captionsetup{labelformat=empty}
        \caption{Gipuma\cite{galliani2015massively}}
    \end{subfigure}

    \begin{subfigure}[t]{\colw\linewidth}
        \begin{minipage}{\textwidth}
            \includegraphics[width=1\linewidth,  keepaspectratio=true,trim={0cm 0cm 0cm 0cm},clip]{{{figures/results/compare/DTU_sparse/model_23/rmvs00_L03}}}
           
        \end{minipage}
        \captionsetup{labelformat=empty}
        \caption{R-MVSNet\cite{yao2019recurrent}}
    \end{subfigure}
    ~ 
    \begin{subfigure}[t]{\colw\linewidth}
        \begin{minipage}{\textwidth}
            \includegraphics[width=1\linewidth,  keepaspectratio=true,trim={0cm 0cm 0cm 0cm},clip]{{{figures/results/compare/DTU_sparse/model_23/pmvs_temper00}}}
           
        \end{minipage}
        \captionsetup{labelformat=empty}
        \caption{Point-MVSNet\cite{chen2019point}}
    \end{subfigure}
    ~ 
    \begin{subfigure}[t]{\colw\linewidth}
        \begin{minipage}{\textwidth}
            \includegraphics[width=1\linewidth,  keepaspectratio=true,trim={0cm 0cm 0cm 0cm},clip]{{{figures/results/compare/DTU_sparse/model_23/colmap00_L03}}}
           
        \end{minipage}
        \captionsetup{labelformat=empty}
        \caption{COLMAP\cite{schoenberger2016mvs}}
    \end{subfigure}

    \caption{Illustration of a very sparse MVS setting using only one seventh of the camera views, \ie $\{v_i\}_{i=1,8,15,22,...}$, to recover the model 23 in the DTU dataset \cite{aanaes2016large}. Compared with the state-of-the-art methods, the proposed SurfaceNet+ provides much complete reconstruction, especially around the boarder region captured by very sparse views.
        } 
   \label{fig:dtu_show}
\end{figure}

We propose an imperative sparse-MVS leader-board and call for the community's attention on the \textit{general} sparse MVS problem with \words{a} large range of baseline angle \words{that could be} up to $~70\degree$. Despite of several approaches recovering 3D model from \words{a} single view, they are biased towards recovering specific objects or scenes with poor generalization ability. For instance, some work focus on improving the depth map generation with the aid of semantic embeddings \cite{chen2016single}\cite{godard2017unsupervised}\cite{liu2015deep} or object-level shape prior \cite{huang2018deep}\cite{mescheder2019occupancy}\cite{kar2017learning}. 
Other methods \cite{schoenberger2016mvs}\cite{paschalidou2018raynet}\cite{yao2019recurrent}\cite{yao2018mvsnet}\cite{chen2019point}\cite{gu2019cascade}, classified as depth map fusion algorithms, try to estimate the depth map for each camera view and fuse them into a 3D model. Unfortunately, for the sparse MVS setting with \words{the} large baseline angle, e.g. \words{larger than} $10 \degree$, these algorithms suffer from incomplete models, because the large baseline angle leads to significantly skewed matching patches from different views and worsens the photo-consistency check. Additionally, as the baseline angle gets larger, the 2D regularization on the depth maps is less helpful for a complete and smooth 3D surface. Because the 2D observation is formed by uneven samples on the 3D surface, the photo consistency agreements can be hardly met by the depth predictions from two views with \words{the} large baseline angle, as shown in Fig.~\ref{fig:dtu_show} and Fig.~\ref{fig:depth_fusion}.
\begin{figure*}[h] 
\centering
    \includegraphics[width=0.95\linewidth]{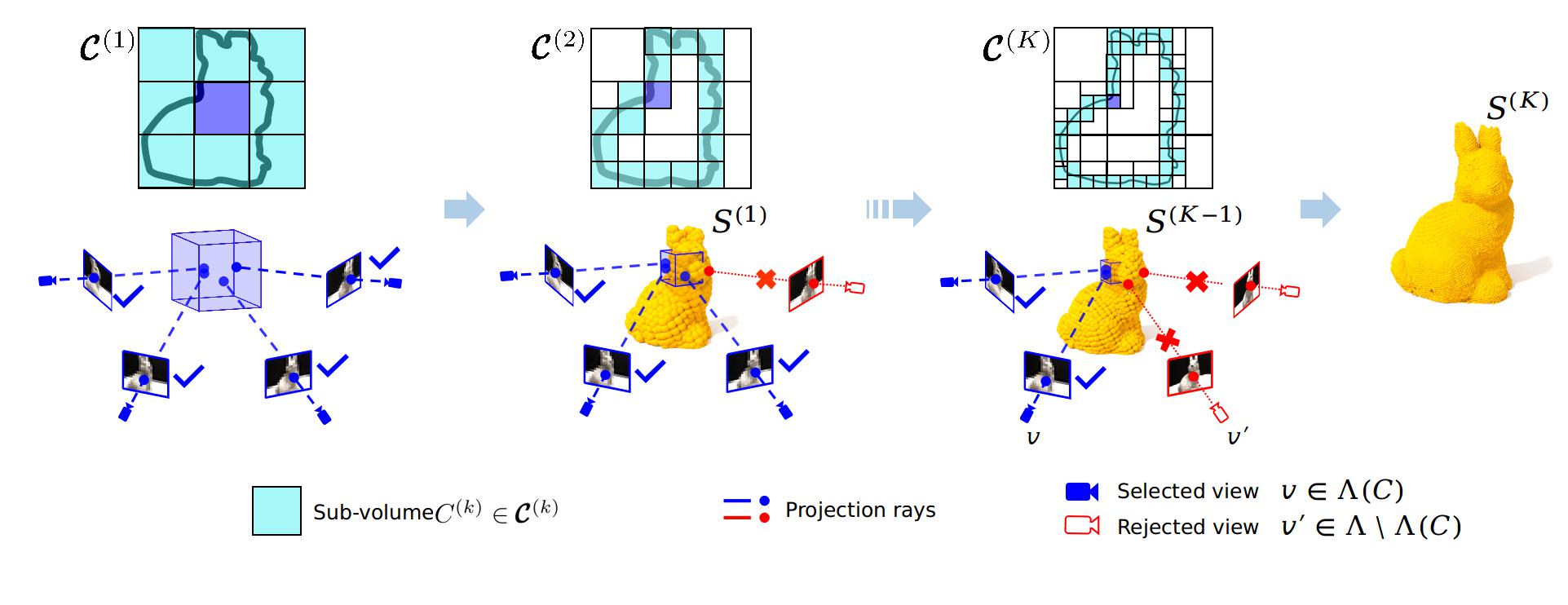} 
    \caption{SurfaceNet+ recovers the whole scene $S^{(K)}$ by progressive refinement of the geometric predictions $S^{(k)}$. So that for each sub-volume $C^{(k)} \in \pmb{\mathcal{C}}^{(k)}$ (drawn as blue cube) the occlusion-aware view selection is performed on the geometric prior. The occluded projection rays are drawn in red and the blue views are the selected ones for reconstruction. In each scale, the volume-wise algorithm only loops through the region in cyan to boost the precision and efficiency.}
    \label{fig:pipeline}
\end{figure*}

\vspace{-0.2cm}
Instead of fusing multiple 2D information into 3D, for the first time, SurfaceNet \cite{ji2017surfacenet} optimizes the 3D geometry in an end-to-end manner by directly learning the volume-wise geometric context from 3D unprojected color volumes. Even though directly utilizing the 3D regularization may avoid the aforementioned shortcomings of the depth map fusion methods, it still suffers from distinct disadvantages such as noisy surface and large holes around the regions with \words{the} repeating pattern and complex geometry. The main reason is that the volume-wise predictions are independently performed without global geometric prior. Consequently, around the region with \words{the} repeating pattern, SurfaceNet returns periodic floating surface fragments around the ground-truth surface. Additionally, such noisy predictions further interferes the view selection and leads to large black holes, as shown in Fig.~\ref{fig:dtu_show}. 

In this paper, we present an end-to-end learning framework, SurfaceNet+, attacking the very sparse MVS problem. \sentences{As the sensation sparsity increase, the number of available photo-consistent views becomes less and the view selection scheme gets more critical.} Therefore, to adapt to \words{a} large range of degree of sparsity, the core innovation is 
\sentences{a trainable occlusion-aware view selection scheme that takes the geometric prior into account via a coarse-to-fine scheme. Such volume-wise view selection strategy can significantly boost the performance of the learning-based volumetric MVS methods.}
More specifically, as shown in Fig.~\ref{fig:pipeline}, it starts from very coarse 3D surface prediction using all the view candidates, and consequently refines the recovered geometry by gradually discarding the occluded views based on the coarser level geometric prediction. Unlike the traditional image-wise \cite{barnes2009patchmatch}\cite{pang2014self}\cite{zheng2015motion} or pixel-wise view selection \cite{schoenberger2016mvs}, which cannot filter out the less irrelevant visible views for the final 3D model fusion, the proposed occlusion-aware volume-wise view selection can identify the most valuable view pairs for each 3D sub-volume and the ranking weights is end-to-end trainable. Therefore, consequently only a little proportion of view pairs is needed for volume-wise surface prediction with little performance reduction. That can dramatically reduce the computational complexity by removing redundancy of the multiview sampling. Benefited from the coarse-to-fine fashion, SurfaceNet+ makes the volume-wise occlusion detection more feasible and leads to a high-recall 3D model.

The proposed sparse-MVS leader-board is built on the large-scale DTU dataset \cite{aanaes2016large} and the Tanks-and-Temples dataset \cite{Knapitsch2017} with sparsely sampled camera views. The sparse-MVS setting selects one view from every $n$ consecutive camera index, \ie $\{1,n+1,2n+1,...\}$, where $n$ is termed as \textit{Sparsity} positively related with baseline angle. The poor performance of the state-of-the-art MVS algorithms on the proposed leader-board demonstrates the necessity of further effort and attention from the community on achieving MVS with various degrees of sparsity. Additionally, the extensive comparison depicts the tremendous performance improvement of SurfaceNet+ over existing methods in terms of precision, recall, and efficiency. As illustrated in Fig.~\ref{fig:dtu_show}, under a very sparse camera setting, SurfaceNet+ predicts a much more complete 3D model compared with recent methods, especially around the border region viewed by \words{a} less number of cameras. \sentences{In summary, the technical contributions in this work are twofold.}
\begin{itemize}
\item \sentences{In consideration of the practical necessity of very sparse MVS and the poor performance of the existing MVS methods, we propose a sparse MVS evaluation benchmark and call for the community's attention on the general sparse MVS problem with a broad range of baseline angles.}
\item \sentences{To tackle with the sparse MVS problem, we propose a novel trainable occlusion-aware view selection scheme, which is a volume-wise strategy and can significantly boost the performance of the volumetric MVS learning framework.
The benchmark and the implementation are publicly available at
\url{https://github.com/mjiUST/SurfaceNet-plus}.
}
\end{itemize}



\section{Related Work}


\subsection{Multi-view  Stereopsis  Reconstruction}
Works in the multi-view stereopsis (MVS) field can be roughly categorised into 1) direct point cloud reconstructions 2) depth maps fusion algorithms and 3) volumetric methods. Point-cloud-based  methods  operate  directly  on  3D  points,  usually  relying  on the propagation strategy to gradually densify the reconstruction \cite{1388267}\cite{5226635}.  As the propagation of point clouds proceeds sequentially, these methods are difficult to be fully parallelized and usually take a long time in the processing. 
Depth maps fusion algorithms\cite{tola2012efficient}\cite{campbell2008using}\cite{galliani2015massively} decouples the complex MVS problem into relatively small problems of per-view depth map estimation, which focus on only one reference and a few source images at a time and then fuse together with the point cloud\cite{MerrellAWMFYNP07}. Yet they suffer from incomplete fusion model with large baseline angle or occluded views since skewed patches and uneven  samples  on  the  3D surface in these cases leads to poor quality
photo consistency agreements.

\words{Volumetric-based methods divide the 3D space into regular grids and handle the problem in a global coordinate. They use either implicit representation\cite{Zach2008}\cite{lempitsky2007global}\cite{curless1996volumetric}\cite{riegler2017octnetfusion} or explicit surface properties\cite{spacecarving99}\cite{Seitz1999}\cite{lsmKarHM2017}\cite{ji2017surfacenet}\cite{jancosek2011multi}\cite{galliani2015massively} to represent and optimize in a global framework. These methods are easy to be parallelized for a multi-view process using a regularization function\cite{lempitsky2007global}\cite{Zach2008} to minimize errors through all points by gradient descent. Though they are more robust to data noise and outliers, the downside of this representation is the high memory consumption, leading to space discretization error, so they are only applicable to synthetic data with low-resolution inputs\cite{lsmKarHM2017}. To deal with the small-scale reconstruction problem, these methods either apply the divide-and-conquer strategy \cite{ji2017surfacenet}, or allow a hierarchical multi-scale structure. \cite{riegler2017octnetfusion}\cite{tatarchenko2017octree} use an octree representation network to represent both the structure of the octree and the probability of each cell and reconstruct the scene in a coarse-to-fine manner, so that time and space complexities are proportional to the size of the reconstructed model. To perceive more geometry details with limited memory, \cite{kazhdan2006poisson}\cite{hane2017hierarchical} adopt a hierarchical adaptive multi-scale algorithm and further facilitates the prediction of high-resolution surfaces. Compared with the mentioned volumetric-based methods, the proposed SurfaceNet+ shares the ideal with the divide-and-conquer strategy but infers the 3D surface in a coarse-to-fine fashion with dynamic view-selection strategy.}



\subsection{Learning-based MVS}


Many learning-based MVS methods have also been developed in recent years.
2D-convolutional neural networks (2D-CNNs)\cite{matchnet_cvpr_15}\cite{Seki2017CVPR}\cite{knoebelreiter_cvpr2017} are applied for better  patch  representation  and  matching, and others such as \cite{7780960} learn the normals of a given depth map to improve the depth map fusion. Yet these methods focus on improving the individual steps in the pipeline and their performance is limited in challenging scenes due to the lack of contextual geometry knowledge. The main promotion in this area is 3D cost volume regularization proposed by \cite{Kendall2017EndtoEndLO}\cite{yao2018mvsnet}\cite{lsmKarHM2017}. This method deploys a 3D volume in the scene space or in the reference camera space. Then, an inverse projection procedure is applied to the 3D volume from several 2D image features gained from different camera positions. Other similar processes such as colored voxel cube\cite{ji2017surfacenet} and recurrent regularization\cite{yao2019recurrent} also use unprojected volumes to get 3D information from 2D image features. The key advantage to process a 3D volume instead of 2D features is that the camera position image information can implicitly write into the 3D volume and the 3D geometry of the scene can be predicted by 3D convolutional layers explicitly. Additionally, during the convolution process, the network is doing the work as in the patch matching method in a highly parallel way, regardless of image distortion and various light conditions.   Our approach  is  more closely  related  to  SurfaceNet\cite{ji2017surfacenet},  which  encodes  camera  geometries  in  the  network  as  multiple unprojected volumes to infer the surface prediction in the global coordinate.

    
\begin{figure}[htbp]
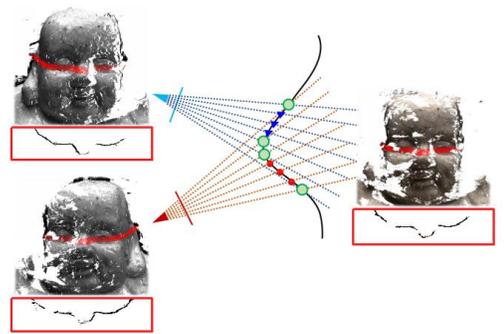
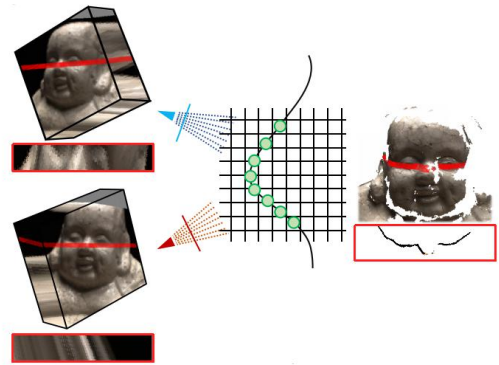
 
    \centering
    \newcommand{\colw}{0.8}
    \newcommand{\figw}{0.8} 
    
     \begin{subfigure}[t]{\colw\linewidth}
        \begin{minipage}{\textwidth}
            \vspace*{0.1cm}
            \includegraphics[width=1\linewidth, keepaspectratio=true,trim={0cm 0cm 0cm 0cm},clip]{{{figures/method/related_depth}}}
           
        \end{minipage}
        \caption{Depth map fusion method. }
        \label{fig:depth_fusion}
    \end{subfigure}
    
     \begin{subfigure}[t]{\colw\linewidth}
        \begin{minipage}{\textwidth}
            \vspace*{0.1cm}
            \includegraphics[width=1\linewidth, keepaspectratio=true,trim={0cm 0cm 0cm 0cm},clip]{{{figures/method/related_volume}}}
           
        \end{minipage}
        \caption{Volumetric method}
        \label{fig:volume_wise}
    \end{subfigure}
    
    \caption{Illustration of two types of  multi-view reconstruction methods. The front view of the 3D model and the top view of the selected region (red) are shown in pair. The circles (green) indicate the prediction. 
    (a): Because the 2D image unevenly samples the 3D surface, as the baseline angle increases, it is rare for view pair (red and blue) to have intersected rays during depth fusion. The 2D regularization gets less helpful.
    (b): Volumetric method optimizes the 3D geometry  by directly regularizing in 3D space.}
    \label{fig:related_discuss}
\end{figure}

\subsection{Depth Map Fusion Methods}\label{depth-fusion-methods}

The depth map fusion algorithms first recover depth maps \cite{xu2013novel} from view pairs by matching similarity patches \cite{barnes2009patchmatch}\cite{pang2014self}\cite{zheng2015motion} along the epipolar line and then fuse the depth maps to obtain a 3D reconstruction of the object \cite{tola2012efficient}\cite{campbell2008using}\cite{galliani2015massively}. \cite{tola2012efficient} is designed for ultra high-resolution image sets and uses a robust descriptor  for efficient matching purposes. In \cite{furu2010accurate} describes a patch model that consists of a quasi-dense set of rectangular patches covering the surface. 
To aggregate image similarity across multiple views, \cite{galliani2015massively} obtains  more accurate depth maps. However, in views with \words{the} large baseline angle it is problematic with the  photo-consistency check because of the significantly skewed patches from different view angles. Therefore, it suffers from incomplete models in sparse-MVS. 

After getting multiple depth maps, the depth map fusion algorithm integrates them into a unified and  augmented scene representation while mitigating any inconsistencies among individual estimates. To improve fusion accuracy, \cite{campbell2008using} learns several sources of the depth map outliers. \cite{Jancosek2011} fuses multiple depth estimations into a surface by evaluating visibility in 3D space, and also attempts to reconstruct the region that is not directly supported by depth measurements. \cite{Goesele2007} proposes to explicitly target the reconstruction from crowd-sourced images. \cite{Zach2008} proposes a variational depth map formulation that enables parallelized computation on the GPU. COLMAP\cite{schoenberger2016mvs} directly maximizes the estimated surface support in the depth maps and allows dataset-wide, pixel-wise sampling for view selection. 
However, as the observations become sparser, 2D depth fusion regularization is less helpful for a complete 3D model, because each 2D view is formed by uneven samples on the 3D surface and the sparse MVS scenario can hardly lead to photo-consistency agreements of the 3D surface prediction from multiple views. \sentences{Compared with the heuristic pixel-wise and image-wise view selection methods that manually filter out the occluded views, the proposed volume-wise view selection method is end-to-end trainable from both geometric and photometric priors for each sub-volume.}

\subsection{Review SurfaceNet}\label{Review of surfaceNet}

SurfaceNet \cite{ji2017surfacenet} firstly proposes an end-to-end learning framework for MVS by automatically learning both photo-consistency and geometric relations of the surface structure. Given two images ($I_i$,$I_j$) and the corresponding camera views ($v_i$,$v_j$), SurfaceNet reconstructs the 3D surface in each sub-volume $C$ by estimating for each voxel $x\in C$ whether it is on the surface or not. 

Firstly, each image of $I_i$ and $I_j$ is unprojected into $C$ by colorizing the voxels on a traced pixel ray into the same pixel color, so that the new representation ($I_i^C$,$I_j^C$) encodes the camera parameters implicitly. 
The gleaming point of the unprojected sub-volume is view-invariant, because the sub-volume is under the global coordinate rather than the relevant coordinate, like the view-variant sweep plane widely used by depth-fusion methods \cite{chen2019point}\cite{yao2019recurrent}. So that it does not lead to the uneven sampling effect.


Then, \words{a pair of} colored voxel cubes ($I_i^C$,$I_j^C$) is fed into SurfaceNet, a fully 3D convolutional neural network, to predict for each voxel $x \in C$ the confidence $p_x \in (0,1)$, which indicates whether a voxel is on the surface or not by using cross-entropy loss. Due to the fully convolutional design, the sub-volume size $s^3$ for inference can be different from that for training, and can be adaptive to various graphic memory sizes.

Lastly, to generalize to a case with multiple views $\Lambda = \{v_1, ..., v_i, ..., v_j, ..., v_V\}$, 
it only selects a subset of view pairs ($v_i,v_j$) to predict $p^{(v_i,v_j)}_x$, \ie the confidence that a voxel $x$ is on the surface, then combines together by taking the weighted average of the predictions
\sentences{based on the relative weight $w_C^{(v_i,v_j)}$ for each view pair 
\begin{equation} \label{eq:old_w}
    w_C^{(v_i,v_j)} = r\left(\theta_C^{(v_i,v_j)}, e(C,I_{v_i}), e(C,I_{v_j}) \right), 
\end{equation}
which is inferred by function $r(\cdot)$ with the inputs of the patch embeddings $e(\cdot)$ and the baseline angle $\theta_C^{(v_i,v_j)}$, i.e., the angle between the projection rays from the center of $C$ to the optical centers of $v_i$ and $v_j$.} So that the volume-wise reconstruction becomes computationally feasible by ignoring the majority of possible view pairs.

Benefited from the direct regularization of the 3D surface, SurfaceNet does not suffer from the shortcoming of 2D regularization owing to the uneven sample of 2D projection. However, the view selection scheme becomes non-trivial and is challenging for the sparse MVS scenario where SurfaceNet still has distinct disadvantages, such as large holes and noisy surfaces around the regions with complex geometry and repeating patterns. Additionally, the volume-wise prediction becomes extremely computationally heavy for large scene reconstruction. In this paper, SurfaceNet+ solves the aforementioned problems with a large margin of performance improvement and around 10X speedup compared with SurfaceNet.

\section{SurfaceNet+}\label{SurfaceNet+}

\begin{figure*}[h] 
    \centering
    \includegraphics[width=0.95\linewidth]{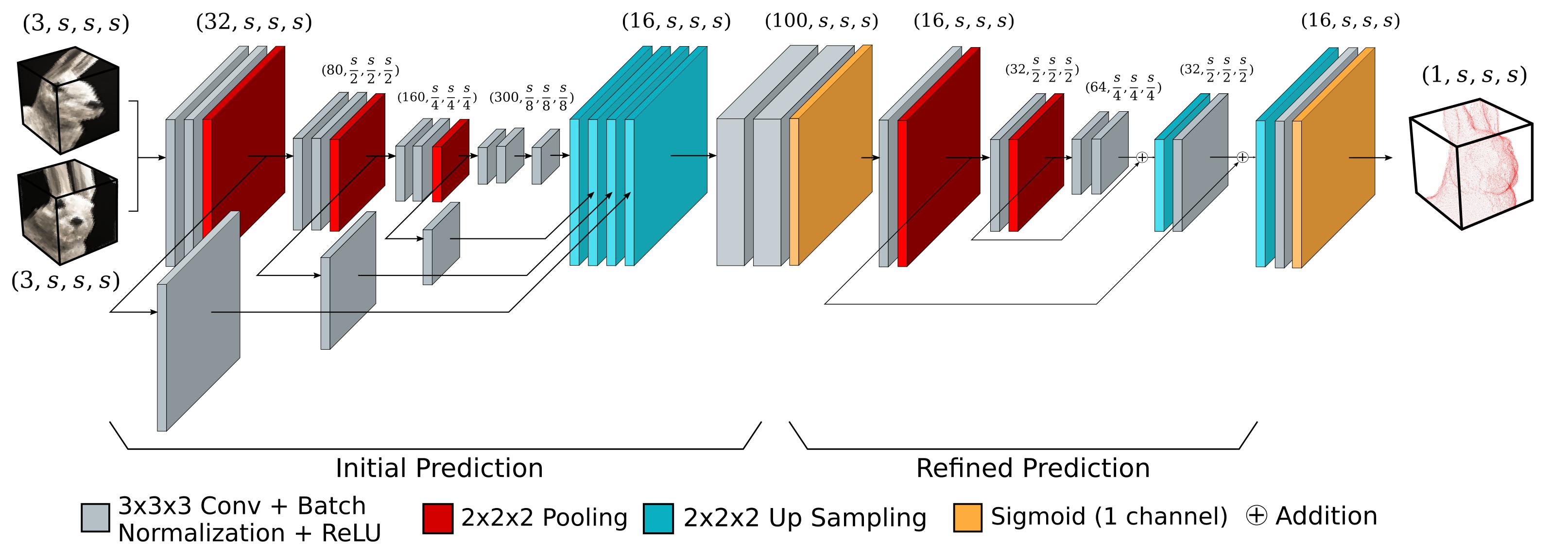} 
    \caption{The network design of SurfaceNet+. The input of the network is two unprojected sub-volumes with size of (3,s,s,s) from different views. The final prediction is an one channel tensor predicting for each voxel the probability of being on surface. }
    \label{fig:network}
\end{figure*}

\sentences{In this Section, We present SurfaceNet+, an end-to-end learning framework, to handle the very sparse MVS problem,
where the critical problem to be solved is the view selection. As the sensation sparsity increases, the number of available photo-consistent views becomes less; thus, the view selection scheme gets more critical. SurfaceNet+ utilizes a novel trainable occlusion-aware view selection scheme that takes the geometric prior into account via a coarse-to-fine strategy. In short, the multi-scale inference (subsection~\ref{inference}) outputs the geometric prior required by the occlusion-aware view selection scheme (subsection~\ref{occlusion-aware}). 
As shown in Fig.~\ref{fig:pipeline}, starting from a bounding box, a very coarse 3D surface is predicted by considering all the view candidates. Subsequently, the coarse level geometry gets iteratively refined by gradually discarding the occluded views based on the coarser level geometric prior.
In subsection~\ref{Network}, the backbone network, a fully convolutional network structure, is presented in detail.}



\subsection{Multi-scale Inference}\label{inference}
For a volume-wise reconstruction pipeline, the noisy prediction occurs frequently around the 3D surface with repeating patterns. Moreover, it suffers from \words{a} huge computational burden to iterate through the majority of the empty space.
While it may be intuitive to consider the 3D geometry prior during reconstruction, the difficulty lies in that the general MVS task does not have any shape prior of the scene. 
What we propose is a coarse-to-fine architecture to gradually refine the geometric details under the assumption that the minority volume of the space is occupied by the 3D surface of the scene.


In the first stage, SurfaceNet+ divides the entire bounding box into a set of sub-volumes $\pmb{\mathcal{C}}^{(1)}$ of the coarsest level with the side length $l^{(1)}=s \cdot r^{(1)}$, where $r^{(1)}$ is the voxel resolution of the coarsest level when the voxelization forms a tensor of size $s \times s \times s$. The tensor size depends highly on the graphic memory size, for example $s=32/64/128$. As the output of this stage, the estimated surface of the coarsest level is denoted as $S^{(1)}$, where $x \in S$ means an occupied voxel in the surface prediction.

\sentences{The following iterative stage divides the space into different sub-volume set in each scale level, \ie $\{\pmb{\mathcal{C}}^{(2)}, \cdots, \pmb{\mathcal{C}}^{(k)}, \cdots, \pmb{\mathcal{C}}^{(K)}\}$, whose resolutions are a geometric sequence with the common ratio $\delta$, \ie $r^{(k)}=\delta \cdot r^{(k+1)}$.} Usually, we set $\delta=4$ to compromise between efficiency and effectiveness. This procedure is iterated until meeting the condition $r^{(K)} \le r$, where $r$ is the desired resolution and $r^{(K)}$ is the finest one. The way to divide the sub-volume is highly dependent on the predicted point cloud of the coarser level $S^{(k-1)}$, when $k=2,3, \cdots$, so that each of the regular sub-volume divisions $C^{(k)} \in \pmb{\mathcal{C}}^{(k)}$ contains at least one point: 
\begin{align} 
    \pmb{\mathcal{C}}^{(k)} = \underset{\pmb{\mathcal{C}}}{\arg\min}\{\left | \pmb{\mathcal{C}} \right | | & \forall \pmb{\mathcal{C}} :  \\
    ( S^{(k-1)}\subseteq \bigcup \pmb{\mathcal{C}} ) \nonumber & \wedge ( \forall C \in \pmb{\mathcal{C}}: S^{(k-1)} \cap C \neq \varnothing ) \}, \label{eq1}
\end{align}
where $| \pmb{\mathcal{C}} |$ denotes the number of sub-volume divisions, and $\bigcup \pmb{\mathcal{C}}$ is a short representation for the union of all the sub-volumes, \ie $\bigcup_{C \in \pmb{\pmb{\mathcal{C}}}}C$. 
To eliminate the boundary effect of the convolution operation, we usually loose the above limitation and allow a slight overlapping between the neighboring sub-volume. The point cloud output $S^{(k)}$ of SurfaceNet+ will be introduced in subsection~\ref{Network}.

\subsection{Trainable Occlusion-aware View Selection}\label{occlusion-aware}
As depicted in Fig.~\ref{fig:dtu_show}, even though SurfaceNet \cite{ji2017surfacenet} does not have the artifacts caused by uneven sampling from 3D surface to 2D depth, it suffers from large holes around the complex geometry. The key reason is that the view selection becomes more critical for the sparse MVS problem. Following the annotation in subsection~\ref{Review of surfaceNet}, \sentences{we introduce how the proposed trainable occlusion-aware view selection scheme can rank and select the top-$N_v$ most valuable view pairs $\pmb{V}_C$ of each sub-volume $C$ from all the possible view pairs
\begin{equation}
    \pmb{V}=\{(v_i, v_j)| (v_i, v_j \in \Lambda) \wedge (v_i \neq v_j)\},
\end{equation}
based on the learned relative weights $w_C^{(v_i,v_j)}$, which is inferred from both the geometric and photometric priors. Note that the multi-scale scheme can provide us with the crucial geometric prior $S^{(k-1)}$. Consequently, according to Eq.~\ref{eq:fusion}, the surface in each sub-volume $C$ is fused by the $|\pmb{V}_C|=N_v$ predictions.} 


\textbf{Geometric Prior.} \sentences{The geometric prior can be easily encoded from the multi-scale predictions. For any camera view $v$ w.r.t. each sub-volume $C \in \pmb{\pmb{\mathcal{C}}}$,} a convex hull $H(C, v) \subset \mathbb{R}^3$ is uniquely defined by a set of points
\begin{equation}
    H(C, v) = Conv(\Gamma(C) \cup \{o_v\}),
\end{equation}
where $o_v$ is the camera center of $v$, and the set $\Gamma(C) = \{c_1, c_2, ..., c_8\}$ contains the 8 corners of $C$.

The more points in the coarser level of surface prediction $S^{(k-1)}$ that appear in the region between the camera view $v$ and the sub-volume $C^{(k)}$, the more likely the view $v$ is occluded. These barrier points are defined in the set
\sentences{
\begin{equation}
    B(C^{(k)}, v) = S^{(k-1)} \cap H(C^{(k)}, v) \textbackslash C^{(k)}.
\end{equation}
}

\textbf{Trainable Relative Weights.} 
\sentences{As suggested in \cite{ji2017surfacenet}, the end-to-end trainable relative weights not only can improve the efficiency by filtering out the majority of the less crucial view pairs for each sub-volume but also can improve the effectiveness of the surface prediction by weighted fusion.
Note that, for sparse MVS, the number of the valid views for each sub-volume could be too few to heuristically detect occlusions. Instead, we propose a trainable occlusion-aware view pair selection scheme that learns the relative weights based on both the geometric and photometric priors:
\begin{equation} \label{eq:new_w}
    w_C^{(v_i,v_j)} = p^{(v_i,v_j)}_{C^{(k)}} \cdot r\left(\theta_C^{(v_i,v_j)}, e(C,I_{i}), e(C,I_{j}) \right),
\end{equation}
where, the photometric priors are the same as SurfaceNet\cite{ji2017surfacenet}, i.e., the baseline angle $\theta_C^{(v_i,v_j)}=\angle o_{v_i} o_C o_{v_j}$ as well as the embeddings $e(\cdot)$ of the cropped patches around the 2D image of $o_C$ on both $I_{i}$ and $I_{j}$ in Eq.~\ref{eq:old_w}, and the geometric prior is encoded as the probability of being not occluded, \ie:\
\begin{align}
    p^{(v_i,v_j)}_{C^{(k)}}= \exp \left(-\alpha \cdot r_{k}^{2} \cdot (|B(C^{(k)}, v_i)| + |B(C^{(k)}, v_j)|) \right), \label{eq:prob}
\end{align}
where $\alpha$ is a hyper parameter controlling the sensitivity of this occlusion probability term and the coefficient $r_{k}^{2}$ can be understood as a normalization term w.r.t. different scales. In Section \label{Quantitative analyze} we will show the effect of $\alpha$ and how it improves the performance of the reconstruction.

\textbf{Weighted Average Surface Prediction.} Lastly, for the general MVS problem, we follow the fusion strategy in SurfaceNet \cite{ji2017surfacenet},
which ranks and selects only a small subset of view pairs $\pmb{V}_C$. Subsequently, the confidence that a voxel $x$ is on the surface, $p_x$, is inferred by the weighted average of the predictions $p^{(v_i,v_j)}_x$:
\begin{equation} \label{eq:fusion}
    p_x = \frac{\sum_{(v_i,v_j) \in \bm{V}_C}w_C^{(v_i,v_j)} p_x^{(v_i, v_j)}}
        {\sum_{(v_i,v_j) \in \bm{V}_C}w_C^{(v_i,v_j)}},
\end{equation}
where $\pmb{V}_C$ denotes the set of selected view pairs with the size of $|\pmb{V}_C|=N_v$, and the relative weight $w_C^{(v_i,v_j)}$ for each view pair is end-to-end trainable and is inferred by Eq.~\ref{eq:new_w}. Note that a smaller $N_v$ can lead to more efficient and less effective results, which is discussed in section~\ref{Ablation}.
}

\subsection{Network}\label{Network}

\textbf{Network Architecture.} At each stage of reconstruction, we use a 3D convolutional neural  network to predict whether each voxel in each sub-volume is on the surface or not. Specifically, given $C^{k}$ and the corresponding image view pairs $(I_{i}, I_{j})$, we first blur each image using a Gaussian kernel to spread the local  information  around  the large receptive field and to guarantee the image consistency in all stages. The unprojected 3D sub-volume $(I_{i}^{C^{(k)}}, I_{j}^{C^{(k)}})$ for a view pair is demonstrated in Fig.~\ref{fig:volume_wise}. The beauty of this representation is that it implicitly encodes the camera parameters as well as scale information to adapt to a fully convolutional neural network.


The detailed network configuration is shown in Fig.~\ref{fig:network}. The building blocks of the model are a UNet-like architecture followed by a refinement network.
SurfaceNet+ takes two colored sub-volumes $(I_{i}^{C^{(k)}}, I_{j}^{C^{(k)}})$ as input, which stores two RGB color values and forms a 6-channel tensor of shape $6 \times s \times s \times s$, and predicts the on-surface probability for each voxel $p_{x \in {C^{(k)}}}$ forming a tensor of size $1 \times s \times s \times s$,
To extract distinct geometry information in various scales, we first use \words{a} pyramid structure to process the features in different receptive fields. To better aggregate multi-scale information, we use two $3\times3$ convolution layers followed by a one-channel convolution layer with a sigmoid activation function after concatenating the features on different scales. 
Inspired by \cite{yao2018mvsnet}, we apply a prediction refinement network at the end of the previous network.  After the initial output $\Tilde{S}_{C}^{(k)}$ with a tensor shape of $1 \times s \times s \times s$, the skip connections at each layer are used to learn the residual prediction and to generate the final output $S_{C}^{(k)}$.



\textbf{Loss.} The training loss consists of two parts to penalize both the initial prediction $p_x$ and the refined prediction $p'_x$. 
In the first stage, the discriminative prediction per voxel $p_x$ is compared with the ground-truth $\hat{s}_x$. 
Since the majority of the voxels does not contain the surface, \ie $\hat{s}_{x \in C^{(k)}}=0$, a class-balanced cross-entropy function is utilized, \ie for each $C^{(k)}$ we have
\begin{align} \label{eq:balancedEntropy}
    & L_{init}= \\
    &\nonumber\;- \sum_{x \in C^{(k)}} \left\{ \beta^{(k)} \hat{s}_x \log p_x + (1-\beta^{(k)})(1-\hat{s}_x)\log (1-p_x) \right\},
\end{align}
where the hyper-parameter $\beta^{(k)}$ is the occupancy ratio of the ground-truth in the scale $k$. 

In the second stage, the refined prediction $p'_x$ is regressed to the ground-truth by the mean square error (MSE), so that the small residue can be penalized as well,
\begin{equation}
L_{refine}=\sum_{x \in C}\left\|\hat{s}_{x} - p'_{x}\right\|_{2},
\end{equation}
where $p_x \in S_{C}^{(k)}$.
Consequently, the training loss is defined as:
\begin{equation}
L_{total}=L_{refine} + L_{init}.
\end{equation}



\begin{figure*}[t]
    \centering
    \newcommand{\colw}{1.0}
    \newcommand{\figw}{1.0} 
    
     \begin{subfigure}[t]{\colw\linewidth}
    \includegraphics[width=\figw\textwidth,trim={0cm 0cm 0cm 0cm},clip]{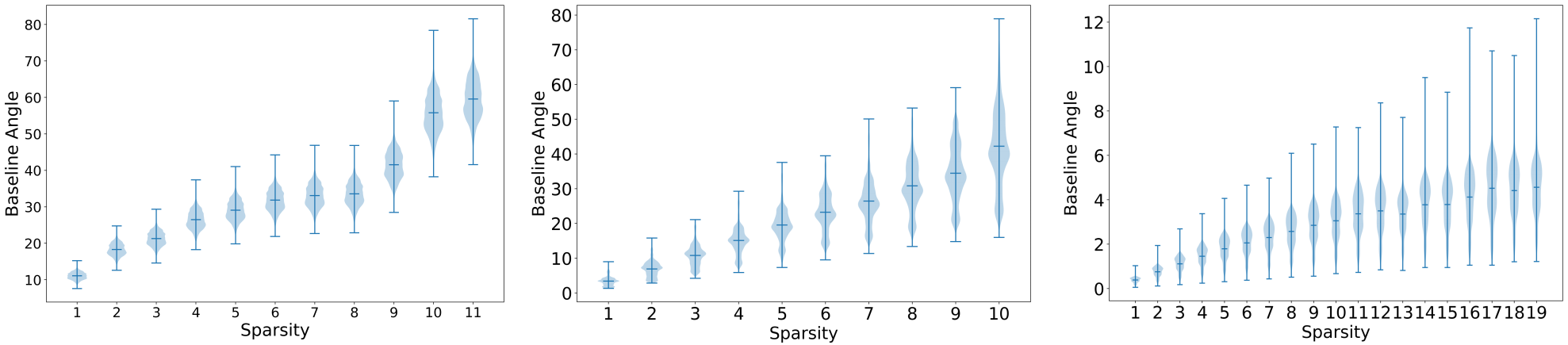}

    \end{subfigure}%
    
    \begin{subfigure}[t]{0.33\textwidth}
        \begin{minipage}{\textwidth}
            \vspace*{0.1cm}
            \caption{DTU\cite{aanaes2016large}}
            
        \end{minipage}
    \end{subfigure}%
    ~
    \begin{subfigure}[t]{0.33\textwidth}
        \begin{minipage}{\textwidth}
            \vspace*{0.1cm}
            \caption{Tanks and Temples Dataset\cite{Knapitsch2017} }
        \end{minipage}
    \end{subfigure}%
    ~
    \begin{subfigure}[t]{0.33\textwidth}
        \begin{minipage}{\textwidth}
            \vspace*{0.1cm}
            \caption{ETH3D low-res dataset\cite{schops2017multi}}
            
        \end{minipage}
    \end{subfigure}%

    \caption{
    \words{The relationship between sparsity and the average baseline angle over all the models in the DTU dataset\cite{aanaes2016large}, the Tanks and Temples dataset\cite{Knapitsch2017} and the ETH3D low-res dataset\cite{schops2017multi}.}
    }
   \label{fig:statistic}
\end{figure*}

\begin{figure*}[h]
    \centering
    \newcommand{\colw}{1.0}
    \newcommand{\figw}{1.0} 
    
     \begin{subfigure}[h]{\colw\linewidth}
    \includegraphics[width=\figw\textwidth,trim={0cm 0cm 0cm 0cm},clip]{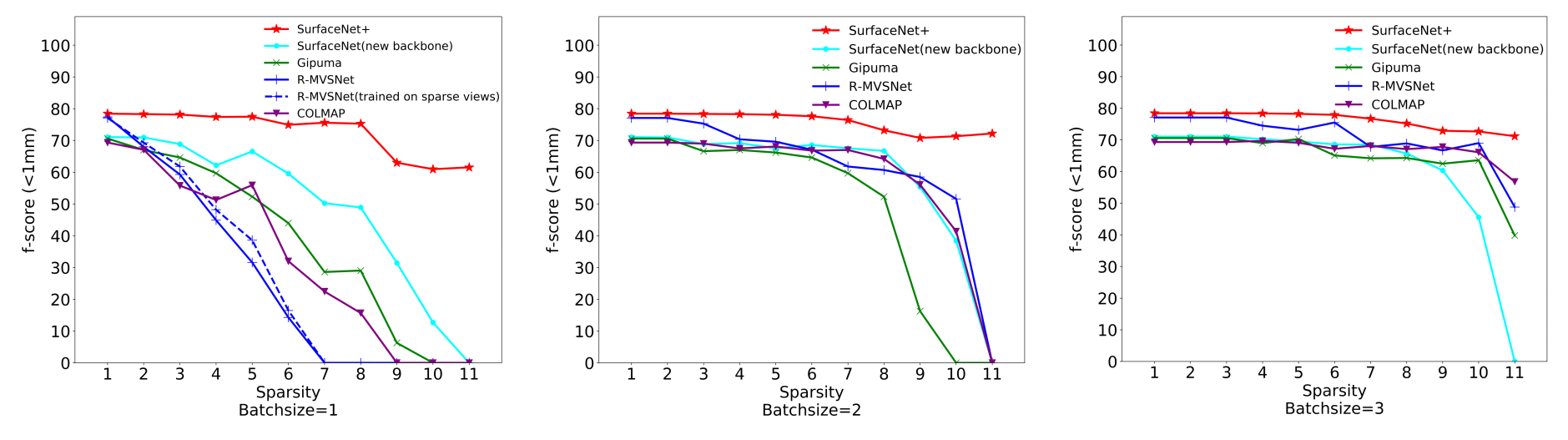}

    \end{subfigure}%
    
            

    \caption{
    \words{\words{Comparison with the existing methods in the DTU Dataset \cite{aanaes2016large} with different sparsely sampling strategy. When $Sparsity=3$ and $Batchsize=2$, the chosen camera indexes are 1,2 $/$ 4,5 $/$ 7,8 $/$ 10,11 $/$ ...}. SurfaceNet+ constantly outperforms the state-of-the-art methods at all the settings, especially at the very sparse scenario.}
    }
   \label{fig:batch_result}
\end{figure*}

\subsection{Implementation Details}



Our network is trained on the DTU dataset \cite{jensen2014large}. We use the volume with $32^3$ voxels to train the network, with a batch size of 16, and the voxel resolution is separately set to 0.4\textit{mm}, 1.0\textit{mm} and 2.0\textit{mm} for each set to generalize on a different scale of surface geometry. To acquire a favorable generalization on sparse-MVS, the network needs to be trained from a variety of view pairs. Therefore, the 3D convolutional network is first trained on randomly-sampled non-occluded view pairs ($v_i,v_j$) without relative weight $w_C^{(v_i,v_j)}$. Then the training process is combined together with $w_C^{(v_i,v_j)}$, and the view pair number is fixed to 6. 
\sentences{Specifically, the relative weights learning procedure is performed using a 2-layer fully connected neural network $r(\cdot)$.}
The computation is introduced in subsection~\ref{occlusion-aware} except that the surface prediction at the previous stage is replaced with the reference model. During the reconstruction stage, the volume size is $64^3$ and the output is upsampled to $128^3$. All the training and reconstruction processes are accomplished on one GTX 1080Ti graphics card.

\section{Sparse-MVS Benchmark}\label{Benchmarks}
In this section, the imperative sparse-MVS leader-board on different datasets, 
the DTU dataset\cite{aanaes2016large}, the Tanks and Temples dataset\cite{Knapitsch2017} (T\&T), and the ETH3D low-res dataset\cite{schops2017multi}, is introduced with extensive comparisons to the recent MVS methods under various observation sparsity levels.

We benchmark SurfaceNet+ at all sparsities from 1 to 11 against several state-of-the-art methods. The sparse MVS setting in our leader-board selects a small proportion of the camera views by consecutively sampling a view $v_i$ from every $sparsity=n$ camera index, \ie $\{1,n+1,2n+1,\cdots\}$.
\sentences{In reality, it is also practical to sample small-batches of images at sparse viewpoints, i.e., grouping batches of views with certain Batchsize at the previously defined sparse viewpoints with a certain Sparsity. When $Sparsity=3$ and $Batchsize=1$, the chosen camera indexes are 1 / 4 / 7 / 10 / $\cdots$. When $Sparsity=3$ and $Batchsize=2$, the chosen camera indexes are 1,2 / 4,5 / 7,8 / 10,11 / $\cdots$.}

Fig.~\ref{fig:statistic} depicts the relationship between sparsity $n$ and the average baseline angle $\bar{\theta}$ averaging over all the ground-truth points in the 22 models of the DTU dataset, 8 models of the Tanks and Temples dataset, and 5 models of the ETH3D low-res dataset, respectively. Note that, for simplicity, only the nearest view pairs are considered to calculate the baseline angle statistics.
\begin{equation}
    \theta = \{\angle o_{v_i}xo_{v_j}| x \in \hat{S}, v_i \in \Lambda, v_j =  \arg\min_{v \in \Lambda}\overline{o_{v_i}o_v}\}
\end{equation}
As the sparsity increases $n=1,...,11$, the average baseline angle $\bar{\theta}$, defined by the intersected projection rays, gradually grows in a large range, e.g. reaching more than $70\degree$ in both DTU and T\&T datasets. Due to the positive correlation between $n$ and $\bar{\theta}$, we claim that our sparse-MVS setting is reasonable by not only covering various degrees of sparsity but also containing irregular sampling locations.

\subsection{DTU Dataset\cite{aanaes2016large}} 
We qualify the performances on the DTU dataset \cite{aanaes2016large} 
in different sparse MVS settings.
The DTU dataset is a large-scale MVS benchmark, which features a variety of objects and materials, and contains 80 different scenes seen from 49 camera positions under seven different lighting conditions. 22 models are selected from the DTU dataset as the evaluation set, following \cite{ji2017surfacenet}
\footnote{\sentences{Follow the same dataset split in SurfaceNet\cite{ji2017surfacenet}.
Training: 
        2, 6, 7, 8, 14, 16, 18, 19, 20, 22, 30, 31, 36, 39, 41, 42, 44, 45, 46, 47, 50, 51, 52, 53, 
        55, 57, 58, 60, 61, 63, 64, 65, 68, 69, 70, 71, 72, 74, 76, 83, 84, 85, 87, 88, 89, 90, 91, 
        92, 93, 94, 95, 96, 97, 98, 99, 100, 101, 102, 103, 104, 105, 107, 108, 109, 111, 112, 
        113, 115, 116, 119, 120, 121, 122, 123, 124, 125, 126, 127, 128. Validation: 3, 5, 17, 21, 28, 35, 37, 38, 40, 43, 56, 59, 66, 67, 82, 86, 106, 117. Evaluation: 1, 4, 9, 10, 11, 12, 13, 15, 23, 24, 29, 32, 33, 34, 48, 49, 62, 75, 77, 110, 114, 118}}.

\begin{figure*}[htbp]
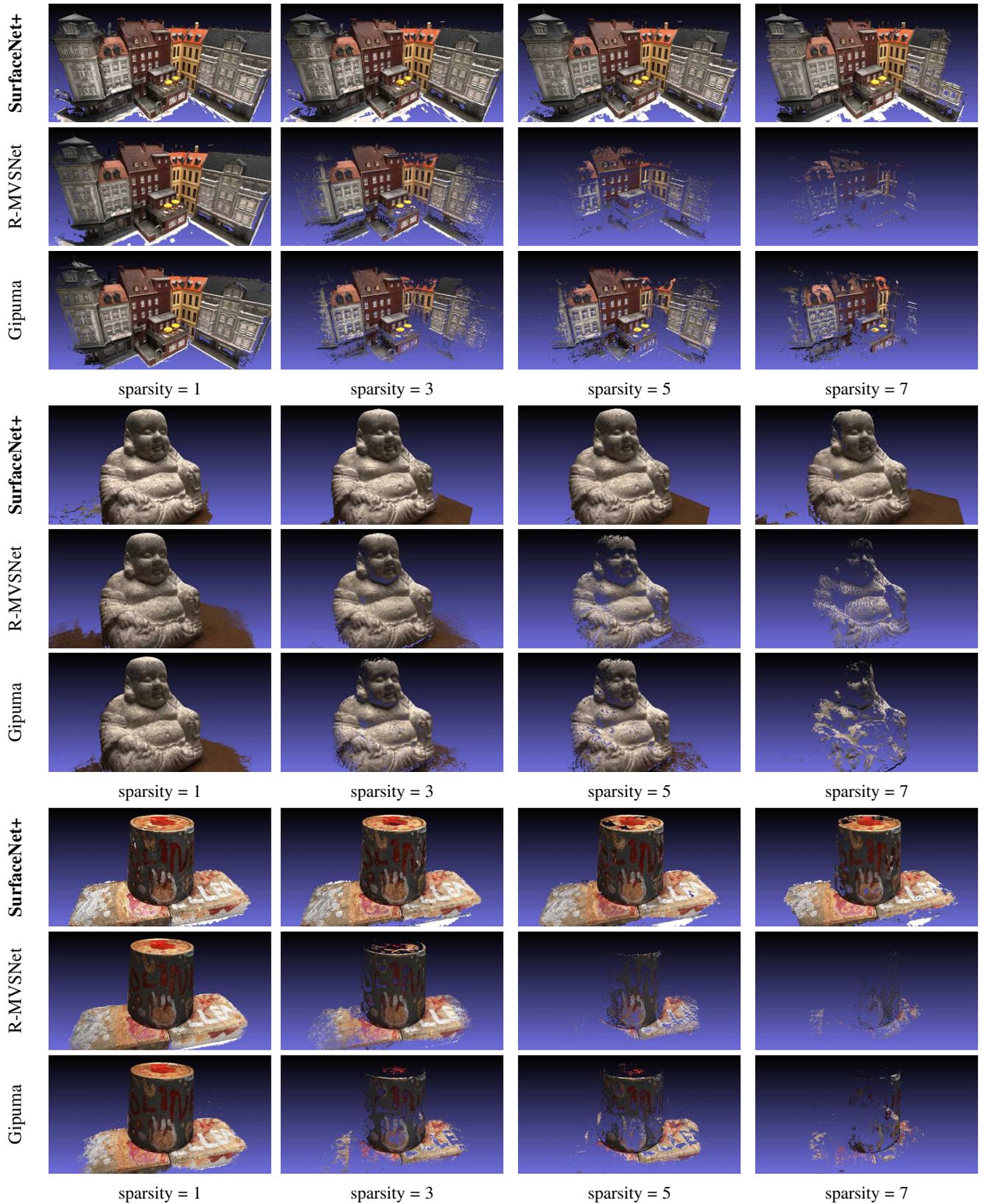

    \centering
    \newcommand{\colw}{0.22}
    \newcommand{\figw}{1.4} 
        
   
   
    \begin{subfigure}[t]{0.03\linewidth}
        \begin{minipage}{\textwidth}
            \vspace*{0.1cm}
            \rotatebox{90}{\textbf{SurfaceNet+}}
        \end{minipage}
    \end{subfigure}
    ~
    \begin{subfigure}[t]{\colw\linewidth}
        \begin{minipage}{\textwidth}
            \vspace*{0.1cm}
            \includegraphics[width=1\linewidth, keepaspectratio=true,trim={0cm 2cm 0cm 0cm},clip]{{{figures/results/compare/DTU_sparse/model_23/surface+00_L00}}}   
            
        \end{minipage}%
    \end{subfigure}%
    ~ 
    \begin{subfigure}[t]{\colw\linewidth}
        \begin{minipage}{\textwidth}
            \vspace*{0.1cm}
            \includegraphics[width=1\linewidth, keepaspectratio=true,trim={0cm 2cm 0cm 0cm},clip]{{{figures/results/compare/DTU_sparse/model_23/surface+00_L01}}}
           
        \end{minipage}
    \end{subfigure}
    ~ 
    \begin{subfigure}[t]{\colw\linewidth}
        \begin{minipage}{\textwidth}
            \vspace*{0.1cm}
            \includegraphics[width=1\linewidth, keepaspectratio=true,trim={0cm 2cm 0cm 0cm},clip]{{{figures/results/compare/DTU_sparse/model_23/surface+00_L02}}}
          
        \end{minipage}
    \end{subfigure}
    ~ 
    \begin{subfigure}[t]{\colw\linewidth}
        \begin{minipage}{\textwidth}
            \vspace*{0.1cm}
            \includegraphics[width=1\linewidth, keepaspectratio=true,trim={0cm 2cm 0cm 0cm},clip]{{{figures/results/compare/DTU_sparse/model_23/surface+00_L03}}}
           
        \end{minipage}
    \end{subfigure}

    \begin{subfigure}[t]{0.03\linewidth}
        \begin{minipage}{\textwidth}
            \vspace*{0.1cm}
            \rotatebox{90}{R-MVSNet}
        \end{minipage}
    \end{subfigure}
    ~
    \begin{subfigure}[t]{\colw\linewidth}
        \begin{minipage}{\textwidth}
            \vspace*{0.1cm}
            \includegraphics[width=1\linewidth, keepaspectratio=true,trim={0cm 2cm 0cm 0cm},clip]{{{figures/results/compare/DTU_sparse/model_23/rmvs00_L00}}}   
            
        \end{minipage}%
    \end{subfigure}%
    ~ 
    \begin{subfigure}[t]{\colw\linewidth}
        \begin{minipage}{\textwidth}
            \vspace*{0.1cm}
            \includegraphics[width=1\linewidth, keepaspectratio=true,trim={0cm 2cm 0cm 0cm},clip]{{{figures/results/compare/DTU_sparse/model_23/rmvs00_L01}}}
           
        \end{minipage}
    \end{subfigure}
    ~ 
    \begin{subfigure}[t]{\colw\linewidth}
        \begin{minipage}{\textwidth}
            \vspace*{0.1cm}
            \includegraphics[width=1\linewidth, keepaspectratio=true,trim={0cm 2cm 0cm 0cm},clip]{{{figures/results/compare/DTU_sparse/model_23/rmvs00_L02}}}
          
        \end{minipage}
    \end{subfigure}
    ~ 
    \begin{subfigure}[t]{\colw\linewidth}
        \begin{minipage}{\textwidth}
            \vspace*{0.1cm}
            \includegraphics[width=1\linewidth, keepaspectratio=true,trim={0cm 2cm 0cm 0cm},clip]{{{figures/results/compare/DTU_sparse/model_23/rmvs00_L03}}}
           
        \end{minipage}
    \end{subfigure}

    \begin{subfigure}[t]{0.03\linewidth}
        \begin{minipage}{\textwidth}
            \vspace*{0.1cm}
            \rotatebox{90}{Gipuma}
        \end{minipage}
    \end{subfigure}%
    ~
    \begin{subfigure}[t]{\colw\linewidth}
        \begin{minipage}{\textwidth}
            \vspace*{0.1cm}
            \includegraphics[width=1\linewidth,  keepaspectratio=true,trim={0cm 2cm 0cm 0cm},clip]{{{figures/results/compare/DTU_sparse/model_23/gipuma00_L00}}}
           
        \end{minipage}
        \captionsetup{labelformat=empty}
        \caption{sparsity = 1}
    \end{subfigure}%
    ~ 
    \begin{subfigure}[t]{\colw\linewidth}
        \begin{minipage}{\textwidth}
            \vspace*{0.1cm}
            \includegraphics[width=1\linewidth,  keepaspectratio=true,trim={0cm 2cm 0cm 0cm},clip]{{{figures/results/compare/DTU_sparse/model_23/gipuma00_L01}}}
           
        \end{minipage}
        \captionsetup{labelformat=empty}
        \caption{sparsity = 3}
    \end{subfigure}
    ~ 
    \begin{subfigure}[t]{\colw\linewidth}
        \begin{minipage}{\textwidth}
            \vspace*{0.1cm}
            \includegraphics[width=1\linewidth,  keepaspectratio=true,trim={0cm 2cm 0cm 0cm},clip]{{{figures/results/compare/DTU_sparse/model_23/gipuma00_L02}}}
           
        \end{minipage}
        \captionsetup{labelformat=empty}
        \caption{sparsity = 5}
    \end{subfigure}
    ~ 
    \begin{subfigure}[t]{\colw\linewidth}
        \begin{minipage}{\textwidth}
            \vspace*{0.1cm}
            \includegraphics[width=1\linewidth,  keepaspectratio=true,trim={0cm 2cm 0cm 0cm},clip]{{{figures/results/compare/DTU_sparse/model_23/gipuma00_L03}}}
           
        \end{minipage}
        \captionsetup{labelformat=empty}
        \caption{sparsity = 7}
    \end{subfigure}
  
    \begin{subfigure}[t]{0.03\linewidth}
        \begin{minipage}{\textwidth}
            \vspace*{0.1cm}
            \rotatebox{90}{\textbf{SurfaceNet+}}
        \end{minipage}
    \end{subfigure}
    ~
    \begin{subfigure}[t]{\colw\linewidth}
        \begin{minipage}{\textwidth}
            \vspace*{0.1cm}
            \includegraphics[width=1\linewidth, keepaspectratio=true,trim={0cm 2cm 0cm 0cm},clip]{{{figures/results/compare/DTU_sparse/model_114/surface+00_L00}}}   
            
        \end{minipage}%
    \end{subfigure}%
    ~ 
    \begin{subfigure}[t]{\colw\linewidth}
        \begin{minipage}{\textwidth}
            \vspace*{0.1cm}
            \includegraphics[width=1\linewidth, keepaspectratio=true,trim={0cm 2cm 0cm 0cm},clip]{{{figures/results/compare/DTU_sparse/model_114/surface+00_L01}}}
           
        \end{minipage}
    \end{subfigure}
    ~ 
    \begin{subfigure}[t]{\colw\linewidth}
        \begin{minipage}{\textwidth}
            \vspace*{0.1cm}
            \includegraphics[width=1\linewidth, keepaspectratio=true,trim={0cm 2cm 0cm 0cm},clip]{{{figures/results/compare/DTU_sparse/model_114/surface+00_L02}}}
          
        \end{minipage}
    \end{subfigure}
    ~ 
    \begin{subfigure}[t]{\colw\linewidth}
        \begin{minipage}{\textwidth}
            \vspace*{0.1cm}
            \includegraphics[width=1\linewidth, keepaspectratio=true,trim={0cm 2cm 0cm 0cm},clip]{{{figures/results/compare/DTU_sparse/model_114/surface+00_L03}}}
           
        \end{minipage}
    \end{subfigure}
   
    \begin{subfigure}[t]{0.03\linewidth}
        \begin{minipage}{\textwidth}
            \vspace*{0.1cm}
            \rotatebox{90}{R-MVSNet}
        \end{minipage}
    \end{subfigure}
    ~
    \begin{subfigure}[t]{\colw\linewidth}
        \begin{minipage}{\textwidth}
            \vspace*{0.1cm}
            \includegraphics[width=1\linewidth, keepaspectratio=true,trim={0cm 2cm 0cm 0cm},clip]{{{figures/results/compare/DTU_sparse/model_114/rmvs00_L00}}}   
            
        \end{minipage}%
    \end{subfigure}%
    ~ 
    \begin{subfigure}[t]{\colw\linewidth}
        \begin{minipage}{\textwidth}
            \vspace*{0.1cm}
            \includegraphics[width=1\linewidth, keepaspectratio=true,trim={0cm 2cm 0cm 0cm},clip]{{{figures/results/compare/DTU_sparse/model_114/rmvs00_L01}}}
           
        \end{minipage}
    \end{subfigure}
    ~ 
    \begin{subfigure}[t]{\colw\linewidth}
        \begin{minipage}{\textwidth}
            \vspace*{0.1cm}
            \includegraphics[width=1\linewidth, keepaspectratio=true,trim={0cm 2cm 0cm 0cm},clip]{{{figures/results/compare/DTU_sparse/model_114/rmvs00_L02}}}
          
        \end{minipage}
    \end{subfigure}
    ~ 
    \begin{subfigure}[t]{\colw\linewidth}
        \begin{minipage}{\textwidth}
            \vspace*{0.1cm}
            \includegraphics[width=1\linewidth, keepaspectratio=true,trim={0cm 2cm 0cm 0cm},clip]{{{figures/results/compare/DTU_sparse/model_114/rmvs00_L03}}}
           
        \end{minipage}
    \end{subfigure}

    \begin{subfigure}[t]{0.03\linewidth}
        \begin{minipage}{\textwidth}
            \vspace*{0.1cm}
            \rotatebox{90}{Gipuma}
        \end{minipage}
    \end{subfigure}%
    ~
    \begin{subfigure}[t]{\colw\linewidth}
        \begin{minipage}{\textwidth}
            \vspace*{0.1cm}
            \includegraphics[width=1\linewidth,  keepaspectratio=true,trim={0cm 2cm 0cm 0cm},clip]{{{figures/results/compare/DTU_sparse/model_114/gipuma00_L00}}}
           
        \end{minipage}
        \captionsetup{labelformat=empty}
        \caption{sparsity = 1}
    \end{subfigure}%
    ~ 
    \begin{subfigure}[t]{\colw\linewidth}
        \begin{minipage}{\textwidth}
            \vspace*{0.1cm}
            \includegraphics[width=1\linewidth,  keepaspectratio=true,trim={0cm 2cm 0cm 0cm},clip]{{{figures/results/compare/DTU_sparse/model_114/gipuma00_L01}}}
           
        \end{minipage}
        \captionsetup{labelformat=empty}
        \caption{sparsity = 3}
    \end{subfigure}
    ~ 
    \begin{subfigure}[t]{\colw\linewidth}
        \begin{minipage}{\textwidth}
            \vspace*{0.1cm}
            \includegraphics[width=1\linewidth,  keepaspectratio=true,trim={0cm 2cm 0cm 0cm},clip]{{{figures/results/compare/DTU_sparse/model_114/gipuma00_L02}}}
           
        \end{minipage}
        \captionsetup{labelformat=empty}
        \caption{sparsity = 5}
    \end{subfigure}
    ~ 
    \begin{subfigure}[t]{\colw\linewidth}
        \begin{minipage}{\textwidth}
            \vspace*{0.1cm}
            \includegraphics[width=1\linewidth,  keepaspectratio=true,trim={0cm 2cm 0cm 0cm},clip]{{{figures/results/compare/DTU_sparse/model_114/gipuma00_L03}}}
           
        \end{minipage}
        \captionsetup{labelformat=empty}
        \caption{sparsity = 7}
    \end{subfigure}

    \begin{subfigure}[t]{0.03\linewidth}
        \begin{minipage}{\textwidth}
            \vspace*{0.1cm}
            \rotatebox{90}{\textbf{SurfaceNet+}}
        \end{minipage}
    \end{subfigure}
    ~
    \begin{subfigure}[t]{\colw\linewidth}
        \begin{minipage}{\textwidth}
            \vspace*{0.1cm}
            \includegraphics[width=1\linewidth, keepaspectratio=true,trim={0cm 2cm 0cm 0cm},clip]{{{figures/results/compare/DTU_sparse/model_1/surface+00_L00}}}   
            
        \end{minipage}%
    \end{subfigure}%
    ~ 
    \begin{subfigure}[t]{\colw\linewidth}
        \begin{minipage}{\textwidth}
            \vspace*{0.1cm}
            \includegraphics[width=1\linewidth, keepaspectratio=true,trim={0cm 2cm 0cm 0cm},clip]{{{figures/results/compare/DTU_sparse/model_1/surface+00_L01}}}
           
        \end{minipage}
    \end{subfigure}
    ~ 
    \begin{subfigure}[t]{\colw\linewidth}
        \begin{minipage}{\textwidth}
            \vspace*{0.1cm}
            \includegraphics[width=1\linewidth, keepaspectratio=true,trim={0cm 2cm 0cm 0cm},clip]{{{figures/results/compare/DTU_sparse/model_1/surface+00_L02}}}
          
        \end{minipage}
    \end{subfigure}
    ~ 
    \begin{subfigure}[t]{\colw\linewidth}
        \begin{minipage}{\textwidth}
            \vspace*{0.1cm}
            \includegraphics[width=1\linewidth, keepaspectratio=true,trim={0cm 2cm 0cm 0cm},clip]{{{figures/results/compare/DTU_sparse/model_1/surface+00_L03}}}
           
        \end{minipage}
    \end{subfigure}

    \begin{subfigure}[t]{0.03\linewidth}
        \begin{minipage}{\textwidth}
            \vspace*{0.1cm}
            \rotatebox{90}{R-MVSNet}
        \end{minipage}
    \end{subfigure}
    ~
    \begin{subfigure}[t]{\colw\linewidth}
        \begin{minipage}{\textwidth}
            \vspace*{0.1cm}
            \includegraphics[width=1\linewidth, keepaspectratio=true,trim={0cm 2cm 0cm 0cm},clip]{{{figures/results/compare/DTU_sparse/model_1/rmvs00_L00}}}   
            
        \end{minipage}%
    \end{subfigure}%
    ~ 
    \begin{subfigure}[t]{\colw\linewidth}
        \begin{minipage}{\textwidth}
            \vspace*{0.1cm}
            \includegraphics[width=1\linewidth, keepaspectratio=true,trim={0cm 2cm 0cm 0cm},clip]{{{figures/results/compare/DTU_sparse/model_1/rmvs00_L01}}}
           
        \end{minipage}
    \end{subfigure}
    ~ 
    \begin{subfigure}[t]{\colw\linewidth}
        \begin{minipage}{\textwidth}
            \vspace*{0.1cm}
            \includegraphics[width=1\linewidth, keepaspectratio=true,trim={0cm 2cm 0cm 0cm},clip]{{{figures/results/compare/DTU_sparse/model_1/rmvs00_L02}}}
          
        \end{minipage}
    \end{subfigure}
    ~ 
    \begin{subfigure}[t]{\colw\linewidth}
        \begin{minipage}{\textwidth}
            \vspace*{0.1cm}
            \includegraphics[width=1\linewidth, keepaspectratio=true,trim={0cm 2cm 0cm 0cm},clip]{{{figures/results/compare/DTU_sparse/model_1/rmvs00_L03}}}
           
        \end{minipage}
    \end{subfigure}

    \begin{subfigure}[t]{0.03\linewidth}
        \begin{minipage}{\textwidth}
            \vspace*{0.1cm}
            \rotatebox{90}{Gipuma}
        \end{minipage}
    \end{subfigure}%
    ~
    \begin{subfigure}[t]{\colw\linewidth}
        \begin{minipage}{\textwidth}
            \vspace*{0.1cm}
            \includegraphics[width=1\linewidth,  keepaspectratio=true,trim={0cm 2cm 0cm 0cm},clip]{{{figures/results/compare/DTU_sparse/model_1/gipuma00_L00}}}
           
        \end{minipage}
        \captionsetup{labelformat=empty}
       \caption{sparsity = 1}
    \end{subfigure}%
    ~ 
    \begin{subfigure}[t]{\colw\linewidth}
        \begin{minipage}{\textwidth}
            \vspace*{0.1cm}
            \includegraphics[width=1\linewidth,  keepaspectratio=true,trim={0cm 2cm 0cm 0cm},clip]{{{figures/results/compare/DTU_sparse/model_1/gipuma00_L01}}}
           
        \end{minipage}
        \captionsetup{labelformat=empty}
        \caption{sparsity = 3}
    \end{subfigure}
    ~ 
    \begin{subfigure}[t]{\colw\linewidth}
        \begin{minipage}{\textwidth}
            \vspace*{0.1cm}
            \includegraphics[width=1\linewidth,  keepaspectratio=true,trim={0cm 2cm 0cm 0cm},clip]{{{figures/results/compare/DTU_sparse/model_1/gipuma00_L02}}}
           
        \end{minipage}
        \captionsetup{labelformat=empty}
        \caption{sparsity = 5}
    \end{subfigure}
    ~ 
    \begin{subfigure}[t]{\colw\linewidth}
        \begin{minipage}{\textwidth}
            \vspace*{0.1cm}
            \includegraphics[width=1\linewidth,  keepaspectratio=true,trim={0cm 2cm 0cm 0cm},clip]{{{figures/results/compare/DTU_sparse/model_1/gipuma00_L03}}}
           
        \end{minipage}
        \captionsetup{labelformat=empty}
        \caption{sparsity = 7}
    \end{subfigure}

    \caption{Quanlitative results of three scans 1, 23 and 114 of the DTU dataset compared with R-MVSNet\cite{yao2019recurrent} and Gipuma\cite{galliani2015massively}. SurfaceNet+
     shows superior performance, particularly with its stable recall quality in sparse cases.  Note that the reconstruction of SurfaceNet+ corresponds to the highest completeness and overall quality as seen in Fig.~ \ref{fig:batch_result}  and Table.~\ref{tab:DTU_score}.
        } 
   \label{fig:dtu_sparse}
\end{figure*}

\begin{figure*}[htbp]
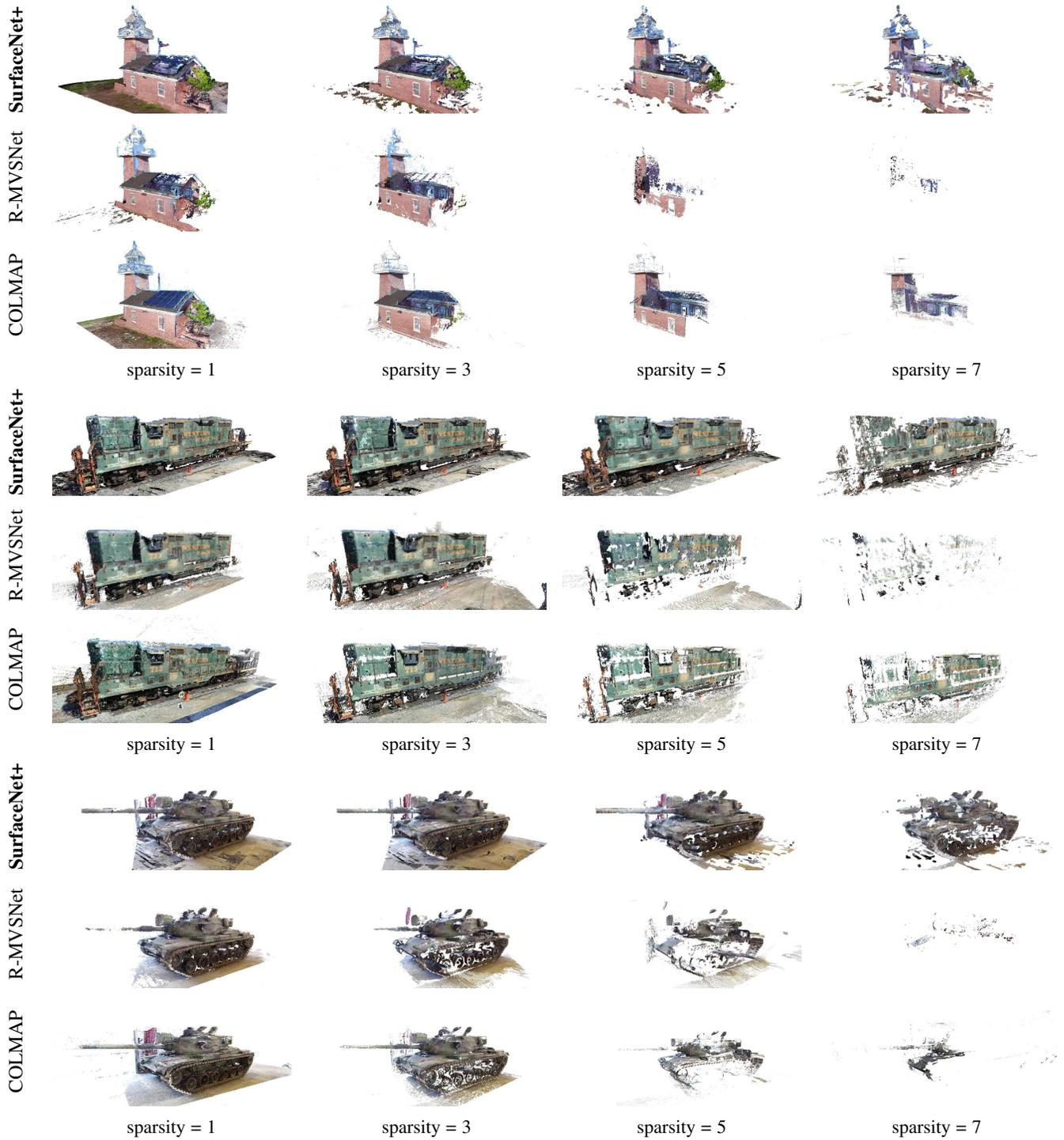

    \centering
    \newcommand{\colw}{0.22}
    \newcommand{\figw}{1.4} 

   
   
    \begin{subfigure}[t]{0.03\linewidth}
        \begin{minipage}{\textwidth}
            \vspace*{0.1cm}
            \rotatebox{90}{\textbf{SurfaceNet+}}
        \end{minipage}
    \end{subfigure}
    ~
    \begin{subfigure}[t]{\colw\linewidth}
        \begin{minipage}{\textwidth}
            \vspace*{0.1cm}
            \includegraphics[width=1\linewidth, keepaspectratio=true,trim={0cm 2cm 0cm 0cm},clip]{{{figures/results/compare/Tanks_sparse/g1/surface+_lighthouse00}}}   
            
        \end{minipage}%
    \end{subfigure}
    ~ 
    \begin{subfigure}[t]{\colw\linewidth}
        \begin{minipage}{\textwidth}
            \vspace*{0.1cm}
            \includegraphics[width=1\linewidth, keepaspectratio=true,trim={0cm 2cm 0cm 0cm},clip]{{{figures/results/compare/Tanks_sparse/g5/surface+_lighthouse00}}}
           
        \end{minipage}
    \end{subfigure}
    ~ 
    \begin{subfigure}[t]{\colw\linewidth}
        \begin{minipage}{\textwidth}
            \vspace*{0.1cm}
            \includegraphics[width=1\linewidth, keepaspectratio=true,trim={0cm 2cm 0cm 0cm},clip]{{{figures/results/compare/Tanks_sparse/g10/surface+_lighthouse00}}}
          
        \end{minipage}
    \end{subfigure}
    ~ 
    \begin{subfigure}[t]{\colw\linewidth}
        \begin{minipage}{\textwidth}
            \vspace*{0.1cm}
            \includegraphics[width=1\linewidth, keepaspectratio=true,trim={0cm 2cm 0cm 0cm},clip]{{{figures/results/compare/Tanks_sparse/g15/surface+_lighthouse00}}}
           
        \end{minipage}
    \end{subfigure}

    \begin{subfigure}[t]{0.03\linewidth}
        \begin{minipage}{\textwidth}
            \vspace*{0.1cm}
            \rotatebox{90}{R-MVSNet}
        \end{minipage}
    \end{subfigure}
    ~
    \begin{subfigure}[t]{\colw\linewidth}
        \begin{minipage}{\textwidth}
            \vspace*{0.1cm}
            \includegraphics[width=1\linewidth, keepaspectratio=true,trim={0cm 2cm 0cm 0cm},clip]{{{figures/results/compare/Tanks_sparse/g1/rmvs_lighthouse00}}}   
            
        \end{minipage}%
    \end{subfigure}
    ~ 
    \begin{subfigure}[t]{\colw\linewidth}
        \begin{minipage}{\textwidth}
            \vspace*{0.1cm}
            \includegraphics[width=1\linewidth, keepaspectratio=true,trim={0cm 2cm 0cm 0cm},clip]{{{figures/results/compare/Tanks_sparse/g5/rmvs_lighthouse00}}}
           
        \end{minipage}
    \end{subfigure}
    ~ 
    \begin{subfigure}[t]{\colw\linewidth}
        \begin{minipage}{\textwidth}
            \vspace*{0.1cm}
            \includegraphics[width=1\linewidth, keepaspectratio=true,trim={0cm 2cm 0cm 0cm},clip]{{{figures/results/compare/Tanks_sparse/g10/rmvs_lighthouse00}}}
          
        \end{minipage}
    \end{subfigure}
    ~ 
    \begin{subfigure}[t]{\colw\linewidth}
        \begin{minipage}{\textwidth}
            \vspace*{0.1cm}
            \includegraphics[width=1\linewidth, keepaspectratio=true,trim={0cm 2cm 0cm 0cm},clip]{{{figures/results/compare/Tanks_sparse/g15/rmvs_lighthouse00}}}
           
        \end{minipage}
    \end{subfigure}

    \begin{subfigure}[b]{0.03\linewidth}
        \begin{minipage}{\textwidth}
            \vspace*{0.1cm}
            \rotatebox{90}{COLMAP}
        \end{minipage}
    \end{subfigure}%
    ~
    \begin{subfigure}[t]{\colw\linewidth}
        \begin{minipage}{\textwidth}
            \vspace*{0.1cm}
            \includegraphics[width=1\linewidth,  keepaspectratio=true,trim={0cm 2cm 0cm 0cm},clip]{{{figures/results/compare/Tanks_sparse/g1/colmap_lighthouse00}}}
           
        \end{minipage}
        \captionsetup{labelformat=empty}
       \caption{sparsity = 1}
    \end{subfigure}
    ~ 
    \begin{subfigure}[t]{\colw\linewidth}
        \begin{minipage}{\textwidth}
            \vspace*{0.1cm}
            \includegraphics[width=1\linewidth,  keepaspectratio=true,trim={0cm 2cm 0cm 0cm},clip]{{{figures/results/compare/Tanks_sparse/g5/colmap_lighthouse00}}}
           
        \end{minipage}
        \captionsetup{labelformat=empty}
        \caption{sparsity = 3}
    \end{subfigure}
    ~ 
    \begin{subfigure}[t]{\colw\linewidth}
        \begin{minipage}{\textwidth}
            \vspace*{0.1cm}
            \includegraphics[width=1\linewidth,  keepaspectratio=true,trim={0cm 2cm 0cm 0cm},clip]{{{figures/results/compare/Tanks_sparse/g10/colmap_lighthouse00}}}
           
        \end{minipage}
        \captionsetup{labelformat=empty}
        \caption{sparsity = 5}
    \end{subfigure}
    ~ 
    \begin{subfigure}[t]{\colw\linewidth}
        \begin{minipage}{\textwidth}
            \vspace*{0.1cm}
            \includegraphics[width=1\linewidth,  keepaspectratio=true,trim={0cm 2cm 0cm 0cm},clip]{{{figures/results/compare/Tanks_sparse/g15/colmap_lighthouse00}}}
           
        \end{minipage}
        \captionsetup{labelformat=empty}
        \caption{sparsity = 7}
    \end{subfigure}

    \begin{subfigure}[t]{0.03\linewidth}
        \begin{minipage}{\textwidth}
            \vspace*{0.1cm}
            \rotatebox{90}{\textbf{SurfaceNet+}}
        \end{minipage}
    \end{subfigure}
    ~
    \begin{subfigure}[t]{\colw\linewidth}
        \begin{minipage}{\textwidth}
            \vspace*{0.1cm}
            \includegraphics[width=1\linewidth, keepaspectratio=true,trim={0cm 2cm 0cm 0cm},clip]{{{figures/results/compare/Tanks_sparse/g1/surface+_train00}}}   
            
        \end{minipage}%
    \end{subfigure}
    ~ 
    \begin{subfigure}[t]{\colw\linewidth}
        \begin{minipage}{\textwidth}
            \vspace*{0.1cm}
            \includegraphics[width=1\linewidth, keepaspectratio=true,trim={0cm 2cm 0cm 0cm},clip]{{{figures/results/compare/Tanks_sparse/g5/surface+_train00}}}
           
        \end{minipage}
    \end{subfigure}
    ~ 
    \begin{subfigure}[t]{\colw\linewidth}
        \begin{minipage}{\textwidth}
            \vspace*{0.1cm}
            \includegraphics[width=1\linewidth, keepaspectratio=true,trim={0cm 2cm 0cm 0cm},clip]{{{figures/results/compare/Tanks_sparse/g10/surface+_train00}}}
          
        \end{minipage}
    \end{subfigure}
    ~ 
    \begin{subfigure}[t]{\colw\linewidth}
        \begin{minipage}{\textwidth}
            \vspace*{0.1cm}
            \includegraphics[width=1\linewidth, keepaspectratio=true,trim={0cm 2cm 0cm 0cm},clip]{{{figures/results/compare/Tanks_sparse/g15/surface+_train00}}}
           
        \end{minipage}
    \end{subfigure}

    \begin{subfigure}[t]{0.03\linewidth}
        \begin{minipage}{\textwidth}
            \vspace*{0.1cm}
            \rotatebox{90}{R-MVSNet}
        \end{minipage}
    \end{subfigure}
    ~
    \begin{subfigure}[t]{\colw\linewidth}
        \begin{minipage}{\textwidth}
            \vspace*{0.1cm}
            \includegraphics[width=1\linewidth, keepaspectratio=true,trim={0cm 2cm 0cm 0cm},clip]{{{figures/results/compare/Tanks_sparse/g1/rmvs_train00}}}   
            
        \end{minipage}%
    \end{subfigure}
    ~ 
    \begin{subfigure}[t]{\colw\linewidth}
        \begin{minipage}{\textwidth}
            \vspace*{0.1cm}
            \includegraphics[width=1\linewidth, keepaspectratio=true,trim={0cm 2cm 0cm 0cm},clip]{{{figures/results/compare/Tanks_sparse/g5/rmvs_train00}}}
           
        \end{minipage}
    \end{subfigure}
    ~ 
    \begin{subfigure}[t]{\colw\linewidth}
        \begin{minipage}{\textwidth}
            \vspace*{0.1cm}
            \includegraphics[width=1\linewidth, keepaspectratio=true,trim={0cm 2cm 0cm 0cm},clip]{{{figures/results/compare/Tanks_sparse/g10/rmvs_train00}}}
          
        \end{minipage}
    \end{subfigure}
    ~ 
    \begin{subfigure}[t]{\colw\linewidth}
        \begin{minipage}{\textwidth}
            \vspace*{0.1cm}
            \includegraphics[width=1\linewidth, keepaspectratio=true,trim={0cm 2cm 0cm 0cm},clip]{{{figures/results/compare/Tanks_sparse/g15/rmvs_train00}}}
           
        \end{minipage}
    \end{subfigure}

    \begin{subfigure}[b]{0.03\linewidth}
        \begin{minipage}{\textwidth}
            \vspace*{0.1cm}
            \rotatebox{90}{COLMAP}
        \end{minipage}
    \end{subfigure}%
    ~
    \begin{subfigure}[t]{\colw\linewidth}
        \begin{minipage}{\textwidth}
            \vspace*{0.1cm}
            \includegraphics[width=1\linewidth,  keepaspectratio=true,trim={0cm 2cm 0cm 0cm},clip]{{{figures/results/compare/Tanks_sparse/g1/colmap_train00}}}
           
        \end{minipage}
        \captionsetup{labelformat=empty}
        \caption{sparsity = 1}
    \end{subfigure}
    ~ 
    \begin{subfigure}[t]{\colw\linewidth}
        \begin{minipage}{\textwidth}
            \vspace*{0.1cm}
            \includegraphics[width=1\linewidth,  keepaspectratio=true,trim={0cm 2cm 0cm 0cm},clip]{{{figures/results/compare/Tanks_sparse/g5/colmap_train00}}}
           
        \end{minipage}
        \captionsetup{labelformat=empty}
        \caption{sparsity = 3}
    \end{subfigure}
    ~ 
    \begin{subfigure}[t]{\colw\linewidth}
        \begin{minipage}{\textwidth}
            \vspace*{0.1cm}
            \includegraphics[width=1\linewidth,  keepaspectratio=true,trim={0cm 2cm 0cm 0cm},clip]{{{figures/results/compare/Tanks_sparse/g10/colmap_train00}}}
           
        \end{minipage}
        \captionsetup{labelformat=empty}
        \caption{sparsity = 5}
    \end{subfigure}
    ~ 
    \begin{subfigure}[t]{\colw\linewidth}
        \begin{minipage}{\textwidth}
            \vspace*{0.1cm}
            \includegraphics[width=1\linewidth,  keepaspectratio=true,trim={0cm 2cm 0cm 0cm},clip]{{{figures/results/compare/Tanks_sparse/g15/colmap_train00}}}
           
        \end{minipage}
        \captionsetup{labelformat=empty}
        \caption{sparsity = 7}
    \end{subfigure}
  
    \begin{subfigure}[t]{0.03\linewidth}
        \begin{minipage}{\textwidth}
            \vspace*{0.1cm}
            \rotatebox{90}{\textbf{SurfaceNet+}}
        \end{minipage}
    \end{subfigure}
    ~
    \begin{subfigure}[t]{\colw\linewidth}
        \begin{minipage}{\textwidth}
            \vspace*{0.1cm}
            \includegraphics[width=1\linewidth, keepaspectratio=true,trim={0cm 2cm 0cm 0cm},clip]{{{figures/results/compare/Tanks_sparse/g1/surface+_m6000}}}   
            
        \end{minipage}%
    \end{subfigure}
    ~ 
    \begin{subfigure}[t]{\colw\linewidth}
        \begin{minipage}{\textwidth}
            \vspace*{0.1cm}
            \includegraphics[width=1\linewidth, keepaspectratio=true,trim={0cm 2cm 0cm 0cm},clip]{{{figures/results/compare/Tanks_sparse/g5/surface+_m6000}}}
           
        \end{minipage}
    \end{subfigure}
    ~ 
    \begin{subfigure}[t]{\colw\linewidth}
        \begin{minipage}{\textwidth}
            \vspace*{0.1cm}
            \includegraphics[width=1\linewidth, keepaspectratio=true,trim={0cm 2cm 0cm 0cm},clip]{{{figures/results/compare/Tanks_sparse/g10/surface+_m6000}}}
          
        \end{minipage}
    \end{subfigure}
    ~ 
    \begin{subfigure}[t]{\colw\linewidth}
        \begin{minipage}{\textwidth}
            \vspace*{0.1cm}
            \includegraphics[width=1\linewidth, keepaspectratio=true,trim={0cm 2cm 0cm 0cm},clip]{{{figures/results/compare/Tanks_sparse/g15/surface+_m6000}}}
           
        \end{minipage}
    \end{subfigure}
   
    \begin{subfigure}[t]{0.03\linewidth}
        \begin{minipage}{\textwidth}
            \vspace*{0.1cm}
            \rotatebox{90}{R-MVSNet}
        \end{minipage}
    \end{subfigure}
    ~
    \begin{subfigure}[t]{\colw\linewidth}
        \begin{minipage}{\textwidth}
            \vspace*{0.1cm}
            \includegraphics[width=1\linewidth, keepaspectratio=true,trim={0cm 2cm 0cm 0cm},clip]{{{figures/results/compare/Tanks_sparse/g1/rmvs_m6000}}}   
            
        \end{minipage}%
    \end{subfigure}
    ~ 
    \begin{subfigure}[t]{\colw\linewidth}
        \begin{minipage}{\textwidth}
            \vspace*{0.1cm}
            \includegraphics[width=1\linewidth, keepaspectratio=true,trim={0cm 2cm 0cm 0cm},clip]{{{figures/results/compare/Tanks_sparse/g5/rmvs_m6000}}}
           
        \end{minipage}
    \end{subfigure}
    ~ 
    \begin{subfigure}[t]{\colw\linewidth}
        \begin{minipage}{\textwidth}
            \vspace*{0.1cm}
            \includegraphics[width=1\linewidth, keepaspectratio=true,trim={0cm 2cm 0cm 0cm},clip]{{{figures/results/compare/Tanks_sparse/g10/rmvs_m6000}}}
          
        \end{minipage}
    \end{subfigure}
    ~ 
    \begin{subfigure}[t]{\colw\linewidth}
        \begin{minipage}{\textwidth}
            \vspace*{0.1cm}
            \includegraphics[width=1\linewidth, keepaspectratio=true,trim={0cm 2cm 0cm 0cm},clip]{{{figures/results/compare/Tanks_sparse/g15/rmvs_m6000}}}
           
        \end{minipage}
    \end{subfigure}

    \begin{subfigure}[b]{0.03\linewidth}
        \begin{minipage}{\textwidth}
            \vspace*{0.1cm}
            \rotatebox{90}{COLMAP}
        \end{minipage}
    \end{subfigure}%
    ~
    \begin{subfigure}[t]{\colw\linewidth}
        \begin{minipage}{\textwidth}
            \vspace*{0.1cm}
            \includegraphics[width=1\linewidth,  keepaspectratio=true,trim={0cm 2cm 0cm 0cm},clip]{{{figures/results/compare/Tanks_sparse/g1/colmap_m6000}}}
           
        \end{minipage}
        \captionsetup{labelformat=empty}
        \caption{sparsity = 1}
    \end{subfigure}
    ~ 
    \begin{subfigure}[t]{\colw\linewidth}
        \begin{minipage}{\textwidth}
            \vspace*{0.1cm}
            \includegraphics[width=1\linewidth,  keepaspectratio=true,trim={0cm 2cm 0cm 0cm},clip]{{{figures/results/compare/Tanks_sparse/g5/colmap_m6000}}}
           
        \end{minipage}
        \captionsetup{labelformat=empty}
        \caption{sparsity = 3}
    \end{subfigure}
    ~ 
    \begin{subfigure}[t]{\colw\linewidth}
        \begin{minipage}{\textwidth}
            \vspace*{0.1cm}
            \includegraphics[width=1\linewidth,  keepaspectratio=true,trim={0cm 2cm 0cm 0cm},clip]{{{figures/results/compare/Tanks_sparse/g10/colmap_m6000}}}
           
        \end{minipage}
        \captionsetup{labelformat=empty}
        \caption{sparsity = 5}
    \end{subfigure}
    ~ 
    \begin{subfigure}[t]{\colw\linewidth}
        \begin{minipage}{\textwidth}
            \vspace*{0.1cm}
            \includegraphics[width=1\linewidth,  keepaspectratio=true,trim={0cm 2cm 0cm 0cm},clip]{{{figures/results/compare/Tanks_sparse/g15/colmap_m6000}}}
           
        \end{minipage}
        \captionsetup{labelformat=empty}
        \caption{sparsity = 7}
    \end{subfigure}

    \caption{Results of three models in Tanks and Temples 'intermediate' set\cite{Knapitsch2017} compared with R-MVSNet\cite{yao2019recurrent} and COLMAP\cite{schoenberger2016mvs}, which demonstrate the power of SurfaceNet+ 
     of high recall prediction in sparse-MVS.  
        } 
   \label{fig:tanks_sparse}
\end{figure*}

\begin{table*}[htbp]
\centering
\tiny
\resizebox{\textwidth}{!}{%
\begin{tabular}{ccccc|ccc|ccc}
\hline
\multirow{2}{*}{Sparsity}                & \multirow{2}{*}{Method}          & \multicolumn{3}{c|}{Mean Distance(mm)}           & \multicolumn{3}{c|}{Percentage(\textless{}1mm)}  & \multicolumn{3}{c}{Percentage(\textless{}2mm)}   \\ \cline{3-11} 
                                         &                                  & Precision      & Recall         & \textbf{Overall}        & Precision      & Recall         & \textbf{f-score}        & Precision      & Recall         & \textbf{f-score}        \\ \hline
\multicolumn{1}{c|}{\multirow{5}{*}{1}}  & \multicolumn{1}{c|}{SurfaceNet+} & 0.385          & \textbf{0.448} & \textbf{0.416} & 88.01          & \textbf{73.01} & \textbf{78.44} & 92.33          & \textbf{78.1}  & \textbf{83.55} \\
\multicolumn{1}{c|}{}                    & \multicolumn{1}{c|}{SurfaceNet\cite{ji2017surfacenet}}  & 0.450          & 1.021          & 0.735          & 84.49          & 64.58          & 71.65          & 89.10          & 68.72          & 76.21          \\
\multicolumn{1}{c|}{}                    & \multicolumn{1}{c|}{Gipuma\cite{galliani2015massively}}      & \textbf{0.283} & 0.873          & 0.578          & \textbf{94.65} & 59.93          & 70.64          & \textbf{96.42} & 63.81          & 74.16          \\
\multicolumn{1}{c|}{}                    & \multicolumn{1}{c|}{R-MVSNet\cite{yao2019recurrent}}    & 0.383          & 0.452          & 0.417          & 87.63          & 72.48          & 77.09          & 91.74          & 76.39          & 82.01          \\
\multicolumn{1}{c|}{}                    & \multicolumn{1}{c|}{COLMAP\cite{schoenberger2016mvs}}      & 0.411          & 0.657         & 0.534          & 82.24          & 52.48          & 61.34          & 88.26          & 62.20          & 72.93          \\ \hline

\multicolumn{1}{c|}{\multirow{5}{*}{3}}  & \multicolumn{1}{c|}{SurfaceNet+} & 0.446          & \textbf{0.482} & \textbf{0.464} & 86.06          & \textbf{74.41} & \textbf{78.15} & 90.87          & \textbf{78.25} & \textbf{82.91} \\
\multicolumn{1}{c|}{}                    & \multicolumn{1}{c|}{SurfaceNet}  & 0.461          & 0.997          & 0.729          & 83.02          & 61.09          & 68.87          & 88.31          & 66.39          & 74.41          \\
\multicolumn{1}{c|}{}                    & \multicolumn{1}{c|}{Gipuma}      & \textbf{0.267} & 1.252          & 0.759          & \textbf{95.51} & 50.88          & 64.63          & \textbf{97.49} & 50.33          & 63.68          \\
\multicolumn{1}{c|}{}                    & \multicolumn{1}{c|}{R-MVSNet}    & 0.465          & 1.012          & 0.738          & 89.55          & 48.03          & 59.28          & 96.96          & 57.92          & 69.04          \\
\multicolumn{1}{c|}{}                    & \multicolumn{1}{c|}{COLMAP}      & 0.467          & 1.090         & 0.778          & 78.45          & 49.26          & 59.62          & 91.44          & 55.98          & 65.77          \\ \hline

\multicolumn{1}{c|}{\multirow{5}{*}{5}}  & \multicolumn{1}{c|}{SurfaceNet+} & 0.446          & \textbf{0.491} & \textbf{0.469} & 88.58          & \textbf{71.63} & \textbf{77.48} & 92.86          & \textbf{76.04} & \textbf{82.28} \\
\multicolumn{1}{c|}{}                    & \multicolumn{1}{c|}{SurfaceNet}  & 0.445          & 0.948          & 0.701          & 81.07          & 58.62          & 66.55          & 85.40          & 62.76          & 70.97          \\
\multicolumn{1}{c|}{}                    & \multicolumn{1}{c|}{Gipuma}      & 0.460          & 1.633          & 1.046          & \textbf{92.38} & 38.53          & 52.36          & \textbf{95.10} & 48.15          & 61.78          \\
\multicolumn{1}{c|}{}                    & \multicolumn{1}{c|}{R-MVSNet}    & \textbf{0.329} & 2.209          & 1.269          & 89.26          & 20.51          & 31.60          & 93.99          & 32.74          & 46.37          \\
\multicolumn{1}{c|}{}                    & \multicolumn{1}{c|}{COLMAP}      & 0.443          & 1.284          & 0.863          & 88.79          & 42.51          & 55.94          & 92.91          & 54.89          & 65.77          \\ \hline

\multicolumn{1}{c|}{\multirow{5}{*}{7}}  & \multicolumn{1}{c|}{SurfaceNet+} & \textbf{0.435} & \textbf{0.524} & \textbf{0.479} & \textbf{91.36} & \textbf{72.23} & \textbf{75.59} & \textbf{95.21} & \textbf{76.54} & \textbf{81.86} \\
\multicolumn{1}{c|}{}                    & \multicolumn{1}{c|}{SurfaceNet}  & 0.688          & 1.130         & 0.909          & 66.86          & 36.91          & 50.24          & 69.21          & 46.91          & 61.70          \\
\multicolumn{1}{c|}{}                    & \multicolumn{1}{c|}{Gipuma}      & 0.569          & 1.770         & 1.169          & 85.35          & 17.91          & 28.66          & 90.78          & 28.00          & 41.31          \\
\multicolumn{1}{c|}{}                    & \multicolumn{1}{c|}{R-MVSNet}    & empty          & empty          & empty          & empty          & empty          & empty          & empty          & empty          & empty          \\
\multicolumn{1}{c|}{}                    & \multicolumn{1}{c|}{COLMAP}      & 0.545          & 1.756          & 1.150          & 59.28          & 15.14          & 22.46          & 80.92          & 31.56          & 41.89          \\ \hline

\multicolumn{1}{c|}{\multirow{5}{*}{9}}  & \multicolumn{1}{c|}{SurfaceNet+} & \textbf{0.441} & \textbf{0.895} & \textbf{0.668} & \textbf{85.99} & \textbf{53.16} & \textbf{63.01} & \textbf{89.86} & \textbf{57.63} & \textbf{67.86} \\
\multicolumn{1}{c|}{}                    & \multicolumn{1}{c|}{SurfaceNet}  & 1.112          & 2.176          & 1.644          & 35.84          & 29.53          & 31.47          & 38.36          & 34.01          & 35.49          \\
\multicolumn{1}{c|}{}                    & \multicolumn{1}{c|}{Gipuma}      & empty          & empty          & empty          & empty          & empty          & empty          & empty          & empty          & empty          \\
\multicolumn{1}{c|}{}                    & \multicolumn{1}{c|}{R-MVSNet}    & empty          & empty          & empty          & empty          & empty          & empty          & empty          & empty          & empty          \\
\multicolumn{1}{c|}{}                    & \multicolumn{1}{c|}{COLMAP}      & empty          & empty          & empty          & empty          & empty          & empty          & empty          & empty          & empty          \\ \hline
\multicolumn{1}{c|}{\multirow{5}{*}{11}} & \multicolumn{1}{c|}{SurfaceNet+} & \textbf{0.445} & \textbf{0.880} & \textbf{0.663} & \textbf{85.81} & \textbf{51.52} & \textbf{61.54} & \textbf{90.05} & \textbf{55.41} & \textbf{65.99} \\
\multicolumn{1}{c|}{}                    & \multicolumn{1}{c|}{SurfaceNet}  & empty          & empty          & empty          & empty          & empty          & empty          & empty          & empty          & empty          \\
\multicolumn{1}{c|}{}                    & \multicolumn{1}{c|}{Gipuma}      & empty          & empty          & empty          & empty          & empty          & empty          & empty          & empty          & empty          \\
\multicolumn{1}{c|}{}                    & \multicolumn{1}{c|}{R-MVSNet}    & empty          & empty          & empty          & empty          & empty          & empty          & empty          & empty          & empty          \\
\multicolumn{1}{c|}{}                    & \multicolumn{1}{c|}{COLMAP}      & empty          & empty          & empty          & empty          & empty          & empty          & empty          & empty          & empty          \\ \hline
\end{tabular}%
}
\captionof{table}{
\words{Quantitative results of reconstruction quality on the DTU dataset in terms of the distance metric(lower is better) and the percentage metric\cite{Knapitsch2017}(higher is better) with 1\textit{mm} and 2\textit{mm} as thresholds. SurfaceNet+ constantly outperforms the state-of-the-arts in all the sparse-MVS settings with $n=1,3,5,7,9,11$.}
} \label{tab:DTU_score}
\end{table*}


\begin{table*}[htbp]
\centering
\resizebox{\textwidth}{!}{%
\begin{tabular}{c|c|ccccccccc}
\hline
 Method                     & \textbf{Average Rank}  & Mean  & Family & Francis & Horse & Lighthouse & M60   & Panther & Playground & Train \\ \hline
 ACMM \cite{xu2019multi}                      & \textbf{14.00}    & 57.27 & 69.24  & 51.45   & 46.97 & 63.20       & 55.07 & 57.64   & 60.08      & 54.48 \\
 CasMVSNet \cite{gu2019cascade}         & \textbf{15.75} & 56.84 & 76.37  & 58.45   & 46.26 & 55.81      & 56.11 & 54.06   & 58.18      & 49.51 \\
 ACMH \cite{xu2019multi}                          & \textbf{22.25} & 54.82 & 69.99  & 49.45   & 45.12 & 59.04      & 52.64 & 52.37   & 58.34      & 51.61 \\
 UCSNet \cite{cheng2019deep}                    & \textbf{22.62} & 54.83 & 76.09  & 53.16   & 43.03 & 54.00         & 55.60  & 51.49   & 57.38      & 47.89 \\
 PLC  \cite{liao2019pyramid}                      & \textbf{24.38} & 54.56 & 70.09  & 50.30    & 41.94 & 58.86      & 49.19 & 55.53   & 56.41      & 54.13 \\
 \textbf{SurfaceNet+}       & \textbf{36.12} & 49.38 & 62.38  & 32.35   & 29.35 & 62.86      & 54.77 & 54.14   & 56.13      & 43.10  \\
 Dense R-MVSNet \cite{yao2019recurrent}           & \textbf{41.00}    & 50.55 & 73.01  & 54.46   & 43.42 & 43.88      & 46.80  & 46.69   & 50.87      & 45.25 \\
 VisibilityAwarePointMVSNet \cite{chen2020visibility} & \textbf{43.88} & 48.70  & 61.95  & 43.73   & 34.45 & 50.01      & 52.67 & 49.71   & 52.29      & 44.75 \\
 Point-MVSNet \cite{ChenPMVSNet2019ICCV}               & \textbf{44.38} & 48.27 & 61.79  & 41.15   & 34.20  & 50.79      & 51.97 & 50.85   & 52.38      & 43.06 \\
 R-MVSNet \cite{yao2019recurrent}                     & \textbf{46.88} & 48.40  & 69.96  & 46.65   & 32.59 & 42.95      & 51.88 & 48.80    & 52.00         & 42.38 \\
 MVSNet \cite{yao2018mvsnet}                     & \textbf{57.50}  & 43.48 & 55.99  & 28.55   & 25.07 & 50.79      & 53.96 & 50.86   & 47.90       & 34.69 \\
 COLMAP \cite{schoenberger2016mvs}                     & \textbf{60.50}  & 42.14 & 50.41  & 22.25   & 25.63 & 56.43      & 44.83 & 46.97   & 48.53      & 42.04 \\ 
\hline
\end{tabular}
}
\captionof{table}{
\words{The top and non-anonymous methods on the Tanks and Temples (T\&T) dataset \cite{Knapitsch2017} leaderboard. The average rank of SurfaceNet+ is higher than that of R-MVSNet \cite{yao2019recurrent}, MVSNet \cite{yao2018mvsnet}, COLMAP \cite{schoenberger2016mvs}, and Point-MVSNet \cite{ChenPMVSNet2019ICCV}.}
} \label{tab:tanks_score}
\end{table*}
\begin{figure*}[t]
    \centering
    \newcommand{\colw}{0.7}
    \newcommand{\figw}{1.0} 
    
     \begin{subfigure}[t]{\colw\linewidth}
    \includegraphics[width=\figw\textwidth,trim={0cm 0cm 0cm 0cm},clip]{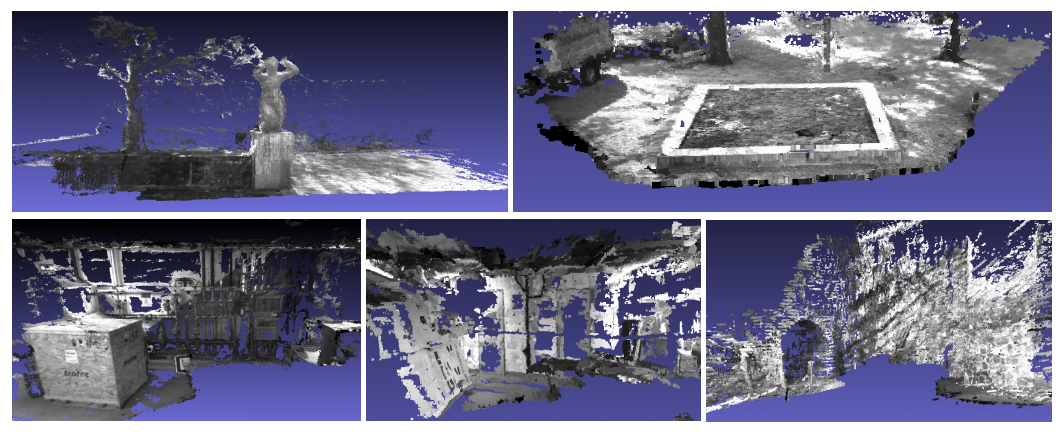}

    \end{subfigure}%
    
            

    \caption{
    \words{Point cloud reconstructions of the ETH3D low-res dataset\cite{schops2017multi}.}
    }
   \label{fig:eth_show}
\end{figure*}

\begin{figure*}[t]
    \centering
    \newcommand{\colw}{0.3}
    \newcommand{\figw}{1.0} 
    
    \begin{subfigure}[t]{\colw\linewidth}
        \includegraphics[width=\figw\textwidth,trim={0cm 0cm 0cm 0cm},clip]{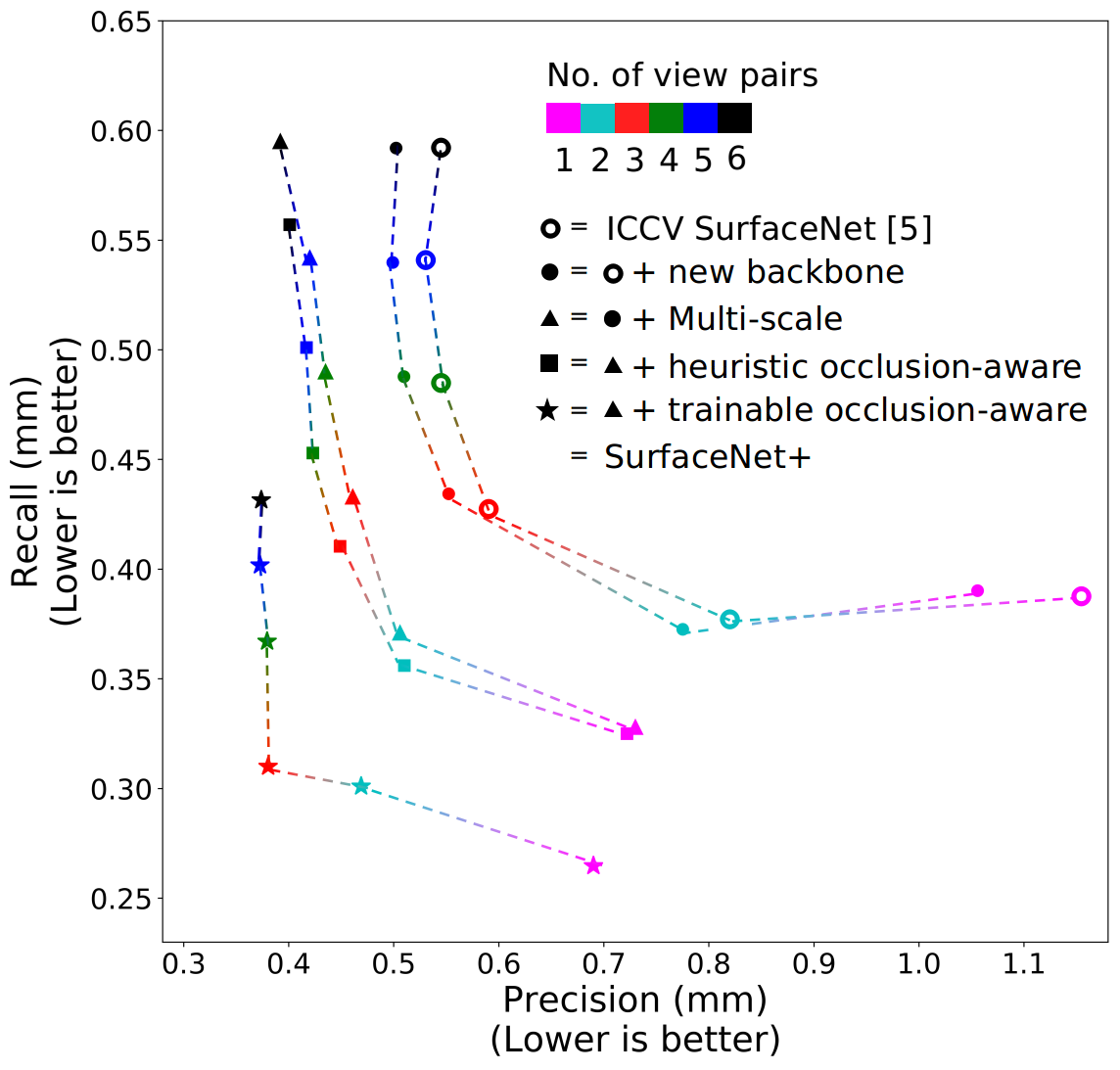}
        \caption{
        \words{
        Ablation study on the ``multi-scale'' and ``occlusion-aware'' modules.}
        }
    \end{subfigure}%
    \label{fig:ablation_picture_a}
    ~ 
    \begin{subfigure}[t]{\colw\linewidth}
        \includegraphics[width=\figw\textwidth,trim={0cm 0cm 0cm 0cm},clip]{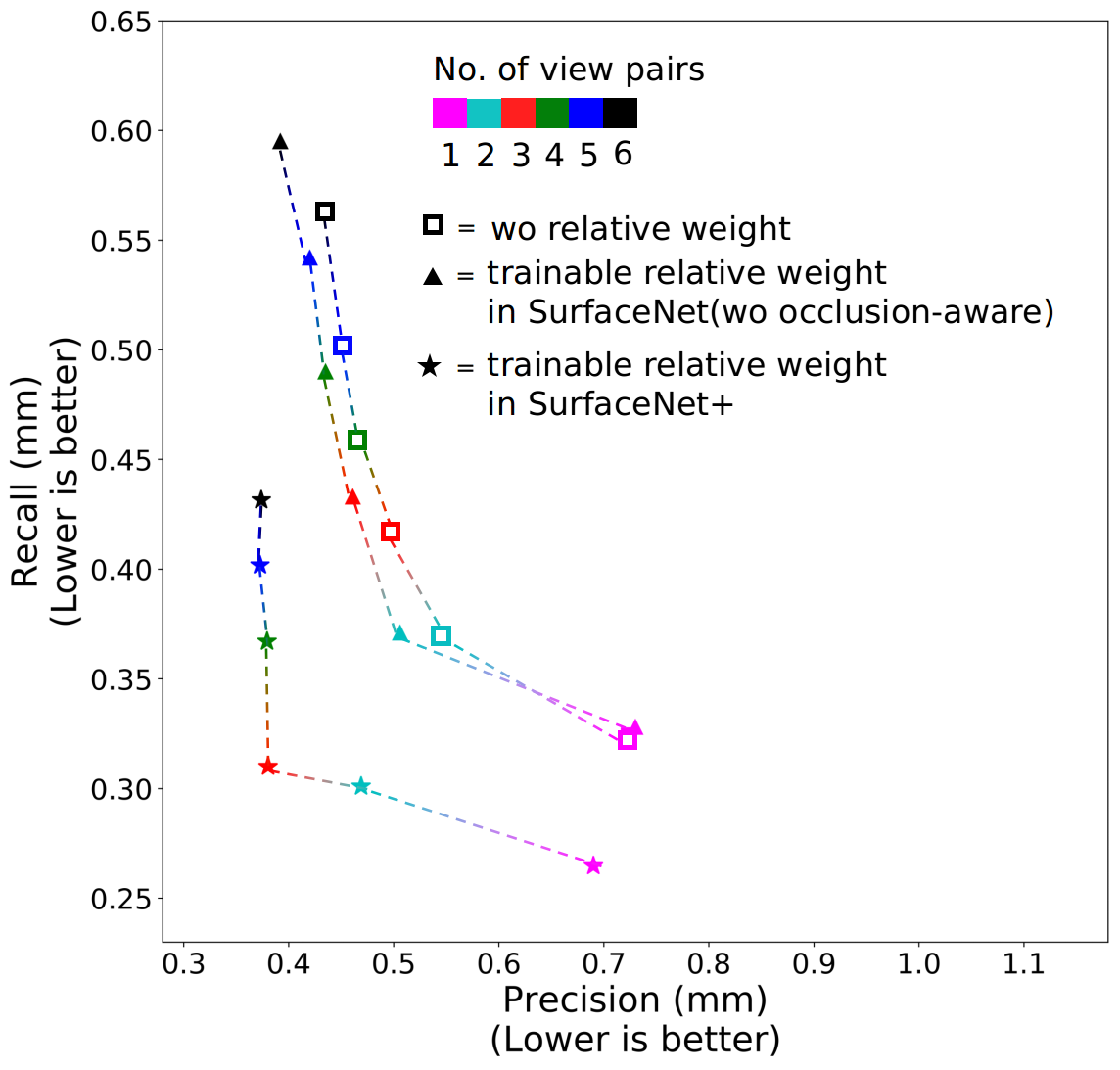}
        \caption{
        \words{Ablation study on the ``relative weight''.}
        }
    \end{subfigure}
    \label{fig:ablation_picture_b}
    ~ 
    \begin{subfigure}[t]{\colw\linewidth}
        \includegraphics[width=\figw\textwidth,trim={0cm 0cm 0cm 0cm},clip]{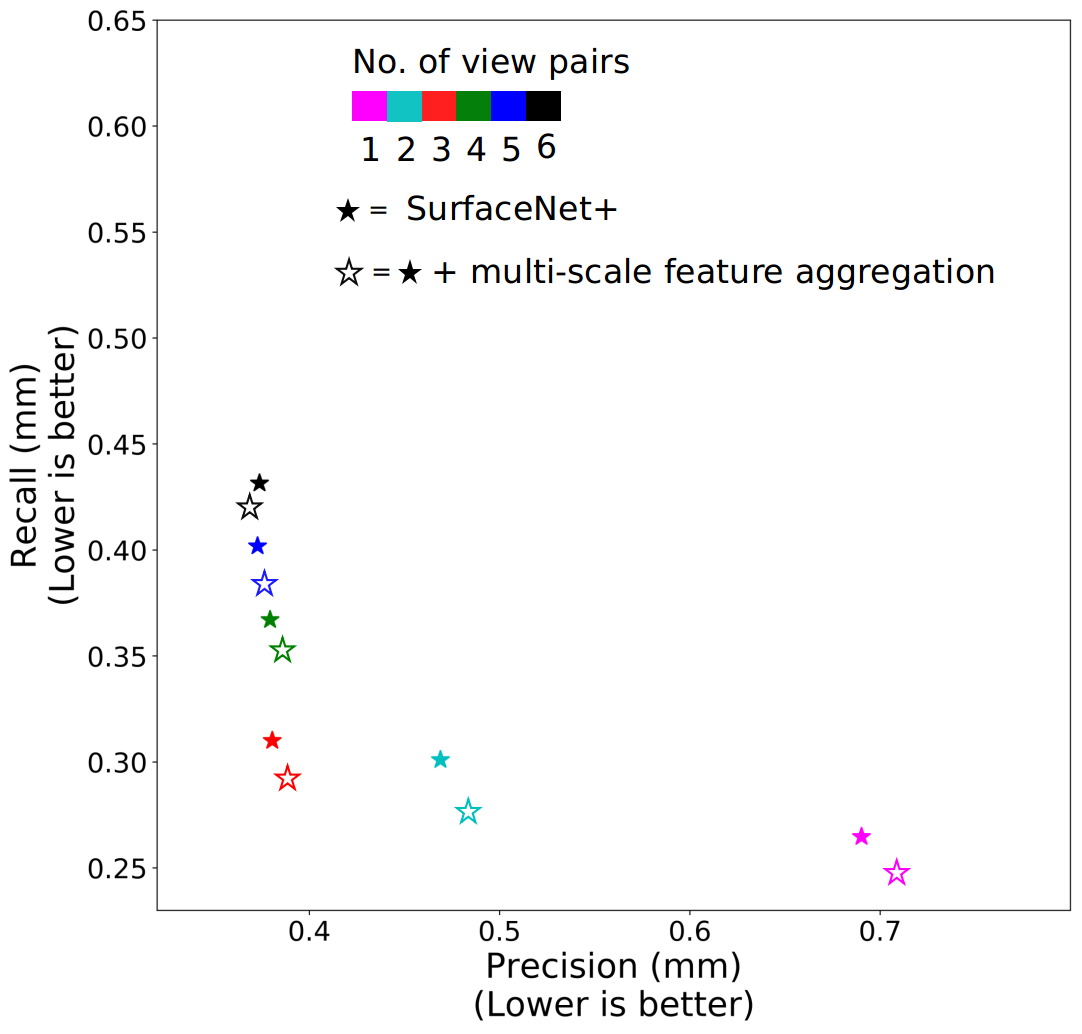}
        \caption{
        \words{Evaluation of a ``multi-scale feature aggregation'' strategy.}
        }
    \end{subfigure}
    \label{fig:ablation_picture_c}
    \caption{
    \words{Ablation study. (a): Comparison among ICCV SurfaceNet \cite{ji2017surfacenet} ($\circ$ curve), SurfaceNet with new backbone ($\bullet$ curve), \textbf{Multi-scale} ($\blacktriangle$ curve), heuristic occlusion-aware view selection during inference ($\blacksquare$ curve) and the proposed trainable occlusion-aware view selection ($\bigstar$ curve). (b): The benefit from an explicit ``relative weight''. $\square$ curve indicates the setting without relative weight that takes heuristic occlusion-aware view selection; $\blacktriangle$ curve is the experiment using trainable relative weight in SurfaceNet (wo occlusion-aware); $\bigstar$ curve depicts the proposed trainable relative weight in SurfaceNet+. (c): Evaluation of the multi-scale feature aggregation strategy that improves the completeness under different number of view pairs.}
    }
   \label{fig:ablation_picture}
\end{figure*}

\begin{figure*}[t]
    \centering
    \newcommand{\colw}{1.0}
    \newcommand{\figw}{1.0} 
    
     \begin{subfigure}[t]{\colw\linewidth}
        \includegraphics[width=\figw\textwidth,trim={0cm 0cm 0cm 0cm},clip]{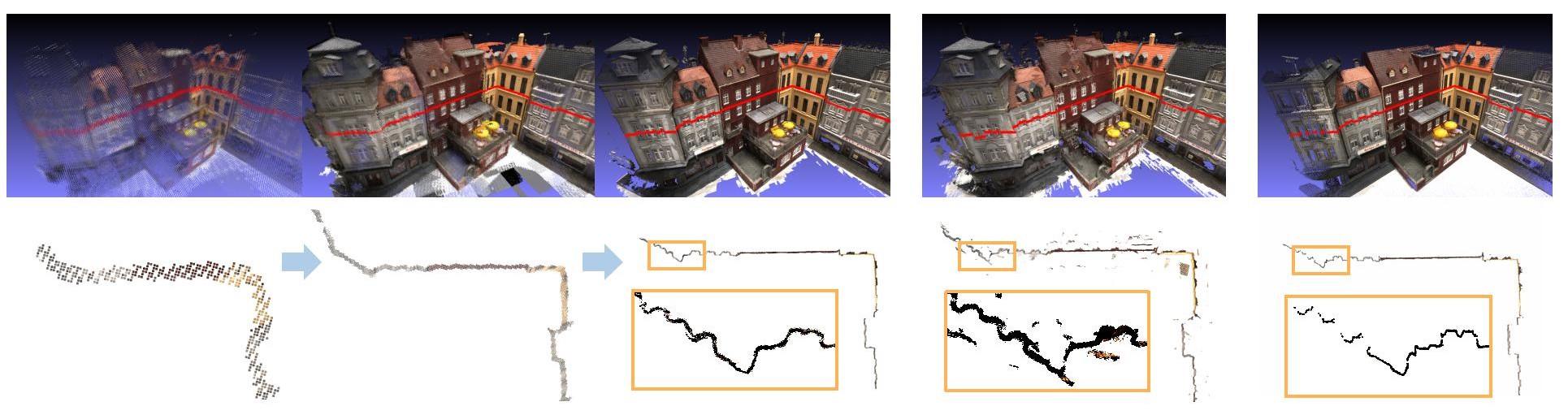}

    \end{subfigure}%
    
    \begin{subfigure}[t]{0.58\textwidth}
        \begin{minipage}{\textwidth}
            \vspace*{0.1cm}
            \caption{\words{Multi-scale results of SurfaceNet+}}
            
        \end{minipage} 
        \label{fig:ablation_volume_a}
        
    \end{subfigure}%
    ~
    \begin{subfigure}[t]{0.21\textwidth}
        \begin{minipage}{\textwidth}
            \vspace*{0.1cm}
            \caption{SurfaceNet}
        \end{minipage}
        \label{fig:ablation_volume-b}
    \end{subfigure}%
    ~
    \begin{subfigure}[t]{0.21\textwidth}
        \begin{minipage}{\textwidth}
            \vspace*{0.1cm}
            \caption{Reference Model}
            
        \end{minipage}
        \label{fig:ablation_volume-c}
    \end{subfigure}%
    \caption{(a): the predictions of three different scales by the coarse-to-fine mechanism, which gradually refines the geometric information. (b): the reconstructed result without the coarse-to-fine mechanism, \ie SurfaceNet \cite{ji2017surfacenet}. (c): the reference model scanned by laser. In each group, we show both (top) the front view of the model and (bottom) the intersection with the red horizontal plane. }
   \label{fig:ablation_volume}
\end{figure*}

\begin{figure*}[htb]
    \centering
    \newcommand{\colw}{0.4}
    \newcommand{\figw}{1.0} 
    \begin{subfigure}[htb]{\colw\linewidth}
        \includegraphics[width=\figw\textwidth,trim={0cm 0cm 0cm 0cm},clip]{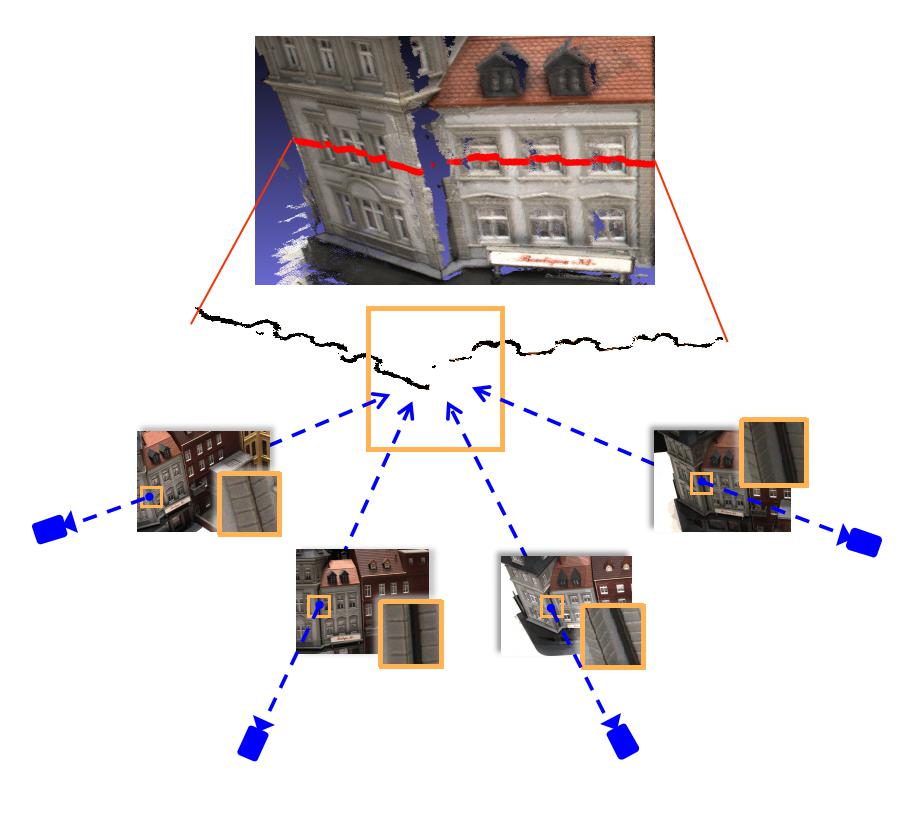}
        \caption{Without occlusion detection}
    \end{subfigure}%
    \hspace{10mm}
    ~ 
    \begin{subfigure}[htb]{\colw\linewidth}
        \includegraphics[width=\figw\textwidth,trim={0cm 0cm 0cm 0cm},clip]{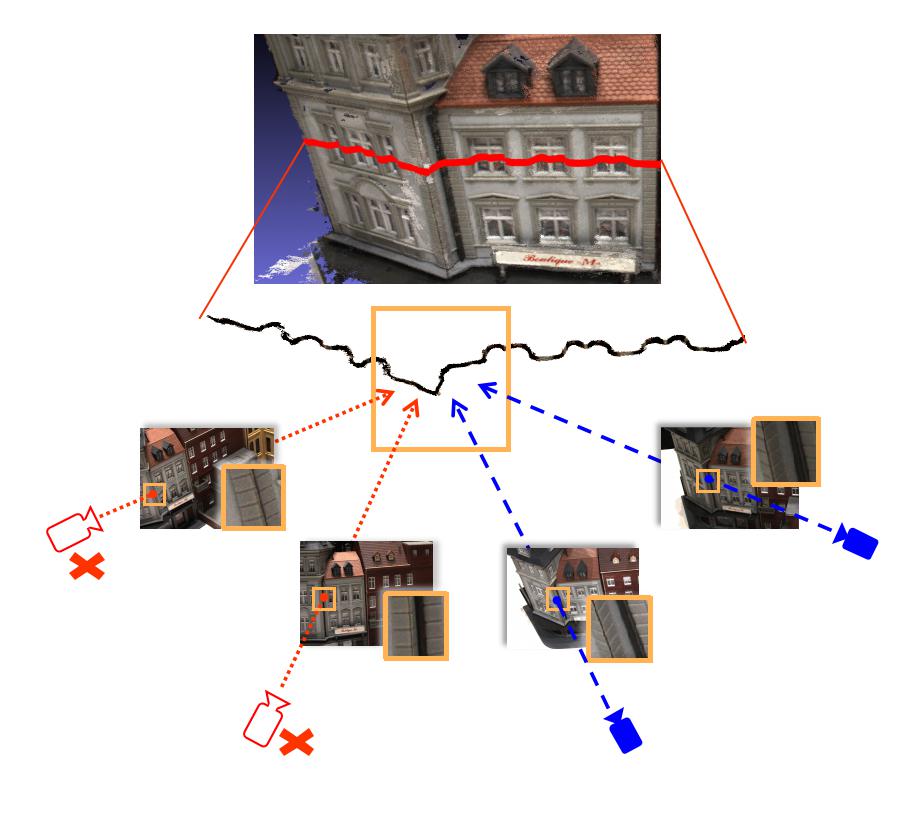}
        \caption{\textbf{SurfaceNet+(with occlusion detection)}}
    \end{subfigure}

    \caption{Qualitative analysis on occlusion detection module. 
    Top: predicted 3D model with selected region (red). 
    Middle: top view of the selected region.
    Bottom: illustration of the selected (red) / rejected (blue) views. (a): the algorithm without occlusion detection leads to large hole around complex geometry, bounded by a yellow square. (b): occlusion-aware view selection is performed by considering geometric prior and significantly improves the recall (completeness).}
   \label{fig:ablation_occlusion}
\end{figure*}

The chart in Fig.~\ref{fig:batch_result} plots the performance under a large range of sparsity in terms of \textit{f-score} (1\textit{mm}), which unifies both \textit{recall} and \textit{precision}. This apparently shows that our proposed method constantly outperforms others in all the sparse settings. Especially for the case of $\bar{\theta} < 40\degree$, amazingly, SurfaceNet+ constantly performs well without obvious degradation. In the extremely sparse case, \ie $\bar{\theta} > 50\degree$, as expected, SurfaceNet+ shows a tiny performance reduction. In contrast, other methods, especially the depth-fusion methods, merely predict a few points. Readers can refer to subsection~\ref{depth-fusion-methods} for the discussion why the depth-fusion methods cannot return a complete result. In our leader-board, depth-map-based methods such as R-MVSNet~\cite{yao2019recurrent} and Gipuma~\cite{galliani2015massively} share the same depth fusion code. For fair comparison, we tuned the hyper-parameters in the depth fusion algorithm to induce better performance in terms of f-score under 1\textit{mm} at each sparsity setting. 

More detailed quantitative results are listed in Table~\ref{tab:DTU_score}, where 3 different matrices are adopted for evaluation.
The precision and recall have two metrics: the distance metric\cite{aanaes2016large} and the percentage metric\cite{Knapitsch2017}. The \textit{overall} score for the percentage metric is measured as the f-score, and a similar measurement for the distance metric overall is given by the average of the mean precision and mean recall.
Obviously, SurfaceNet+ outperforms the state-of-the-art methods in both recall and precision at all sparsity settings. Unlike other methods whose recall dramatically decay when the sparsity increases, SurfaceNet+ has almost consistent recall quality with high precision.

The qualitative comparison of SurfaceNet+ and the other two methods, R-MVSNet~\cite{yao2019recurrent} and Gipuma~\cite{galliani2015massively}, is illustrated in Fig.~\ref{fig:dtu_sparse}, showing that SurfaceNet+ precisely reconstructs the scenes while maintaining high recall. In particular, SurfaceNet+ is able to generate a high-recall point cloud in complex geometry and repeating pattern regions when $sparsity=7$, which means it evenly fuses the accurate 3D model with corrected-selected non-occluded views. The detailed analysis is shown in Section \ref{Ablation}. 

\words{
To have a slightly different way of sparse sampling, three $Batchsize$ values $\{1,2,3\}$ are evaluated as depicted in Fig.~\ref{fig:batch_result}. It can be observed that SurfaceNet+ constantly outperforms others despite that the depth-fusion methods (Gipuma\cite{galliani2015massively}, R-MVSNet\cite{yao2019recurrent}, COLMAP\cite{schoenberger2016mvs}) boost the performance as the $Batchsize$ increases. Moreover, as the disparity increases, the performance drop of the existing methods is apparently larger than that of SurfaceNet+.
In particular, we have retrained the R-MVSNet for sparse MVS with randomly-sampled non-occluded view pairs at $Batchsize=1$. As shown in Fig.~\ref{fig:batch_result}, the gain is inapparent in terms of f-score. As the depth-fusion based MVS methods (R-MVSNet) rely more on the photo-consistency in 2D images, the large baseline angles of a very sparse MVS problem leads to severely skewed matching patches across views that significantly toughen the dense correspondence problem. In contrast, the learning-based volumetric MVS methods like SurfaceNet+ avoids the 2D correspondence search problem by directly inferring 3D surface from each unprojected 3D sub-volumes. That may explain why the learning-based volumetric methods outperform the depth-fusion based methods in the very sparse MVS settings.
}
\words{
For the experiment settings, both R-MVSNet and Gipuma shared the same depth fusion code, and we tuned the hyper-parameters of it to induce better performance in terms of f-score under 1mm at each sparsity setting. More specifically, followed by Gipuma\cite{galliani2015massively}, since there is a tradeoff between accuracy and completeness, we choose the depth fusion parameter settings that achieve high accuracy at sparsity=1,2 and high completeness at sparsity>=3. The other part remain the same as the paper of R-MVSNet\cite{yao2019recurrent} and Gipuma\cite{galliani2015massively}. In COLMAP\cite{schoenberger2016mvs}, all parameters were set as the default values.
}


\subsection{Tanks and Temples Dataset\cite{Knapitsch2017}} 

The Tanks and Temples (T$\&$T) dataset\cite{Knapitsch2017} contains real-world large scenes under complex lighting conditions. In Fig.~\ref{fig:tanks_sparse}, we compare the qualitative results in the Tanks and Temples dataset\cite{Knapitsch2017} with R-MVSNet~\cite{yao2019recurrent} and COLMAP~\cite{schoenberger2016mvs}. The results indicate the effectiveness of our proposed method at different sparsities. \words{We submitted and evaluated the SurfaceNet+ results ($Sparsity=1$) to the online leader-board. As depicted in Table~\ref{tab:tanks_score}, despite the dense MVS condition, the overall rank of SurfaceNet+ is still higher than that of R-MVSNet\cite{yao2019recurrent}, MVSNet\cite{yao2018mvsnet}, COLMAP\cite{schoenberger2016mvs}, and Point-MVSNet\cite{ChenPMVSNet2019ICCV}. Note that we list and compare with all the top and non-anonymous methods on the leaderboard in the following table.}


\subsection{Generalization on the ETH3D Dataset\cite{schops2017multi}}

\words{We also evaluate the generalization ability by adopting the ETH3D dataset\cite{schops2017multi}, \ie we direct evaluate the proposed method that trained only on the DTU training dataset without fine-tuning the network. The results of the low-resolution scenes are shown in Fig.~\ref{fig:eth_show}. It is worth noting that the baseline angle in the ETH3D dataset is tiny among all the camera views because the images were acquired by just rotating the camera with little camera translation. Fig.~\ref{fig:statistic}(c) further depicts the relationship between sparsity and the average baseline angle over all the models in the ETH3D low-resolution training set. The average baseline angle is far less than $~8\degree$, indicating that the ETH3D dataset may not be suitable for the sparse-MVS benchmark.}

\section{Ablation Study} \label{Ablation}
To investigate the influences of each of the key components in the proposed method, we design an ablation study with respective to the coarse-to-fine fashion (\textbf{Multi-scale}) and the  
\words{trainable occlusion-aware view selection (\textbf{View-selection})}.
For all these studies below, experiments are performed and evaluated on a specific model (model 23) in the DTU dataset because it contains many challenging cases such as complex geometry, textureless regions, and repeating patterns.

In the sparse case, for example, $sparsity=5$, we quantitatively illustrate the performance gain of the multi-scale fashion in Fig.~\ref{fig:ablation_picture}(a), \words{in which we compare few settings: ICCV SurfaceNet \cite{ji2017surfacenet} ($\circ$ curve), ICCV SurfaceNet with the new backbone ($\bullet$ curve) denoted as SurfaceNet in the rest of the paper, SurfaceNet with multi-scale inference ($\blacktriangle$ curve), and the proposed trainable occlusion-aware view selection scheme ($\bigstar$ curve).
Clearly, from the comparison of $\blacktriangle$ v.s. $\bigstar$, we can conclude that the proposed trainable occlusion-aware view selection scheme that is a volume-wise strategy significantly improves both completeness (Recall) and accuracy (Precision).  
}

\subsection{Multi-scale Mechanism} 
 Fig.~\ref{fig:ablation_volume}(a) shows the predictions of the various scale levels. Note that the  volume-wise occlusion detection is turned off. Fig.~\ref{fig:ablation_volume}(b) contains the result without using the coarse-to-fine mechanism, which is the same as SurfaceNet\cite{ji2017surfacenet}. The reference model scanned by laser is placed in Fig.~\ref{fig:ablation_volume}(c). In each group, (top) the front view of the reconstruction model and (bottom) the intersection of a horizontal plane (red line) are shown. The top view of the red line is useful to observe the surface thickness, noise level, and completeness.

 
 Comparing (a) and (b), it is obvious that the method with the coarse-to-fine mechanism leads to higher precision at the texture area and complex geometry region. Although (b) accurately predicts the results at some complex regions, it suffers from thick surface prediction and floating noise around the repeating pattern regions. The floating noise occurs close to the real surface, because the volume-wise method processes each sub-volume locally and individually without global prior to filter out the floating noise. In contrast, the coarse-to-fine mechanism is helpful to gradually reject the empty space and to refine the geometric prediction.

 In the sparse case, for example, $n=5$, the multi-scale mechanism dramatically improves precision if we compare the round-curve and the triangle-curve in Fig.~\ref{fig:ablation_picture}(a). Apparently the triangle-curve is a shifted version of the round-curve towards the direction for better precision with constant recall.
 
\words{\textbf{Feature aggregation.} To give the network more global context, we try to use some features coming from the coarser level of the network so that the coarse level is used to not only decide on the visibility/occlusions, but also provide additional feature contents. We study the advantages and disadvantages of this multi-scale inference architectures and report the results in Fig.~\ref{fig:ablation_picture}(c). It can be shown that the multi-scale feature aggregation scheme (\FiveStarOpen) improves the completeness (Recall) of the results by providing the global context. However, when there are few numbers of view pairs, e.g., less than 6 view pairs, the multi-scale aggregation worsens the accuracy (Precision) of the prediction. The reason is that the volume-wise surface prediction relies on multiple pairs of the unprojected sub-volumes, and in the coarse-to-fine procedure, the selected views may be updated based on the geometric priors under different scales. So that when the multi-scale scheme aggregates the features from different view pairs, the global context may become less useful and leads to worse accuracy (Precision).
}


\subsection{Occlusion Detection}
To analyze the qualitative impact of occlusion-aware view selection, the comparison experiment is set based on the result using the multi-scale mechanism. For better visualization, we only probe the occlusion-aware view selection in the final multi-scale stage. The results with and without occlusion-aware view selection are shown in Fig.~\ref{fig:ablation_occlusion}, which contains the front views and the intersection (the red line shown in the model) of the results accompanied by different camera views.


Note that SurfaceNet+ (with View-selection) has higher recall output, especially around the corner of the reconstructed house model (shown in the orange box of the intersection). The gap lies in different views selected by each method. Both methods use patch image (bottom right corner of the picture) to select valid views (the four views shown in the bottom of the figure).
Yet the left two views are blocked by the surface, which means only the right two views can provide useful patch information for reconstruction. The occluded views reduced the output weight under the correct views; therefore, incomplete prediction occurred in complex geometry regions without occlusion-aware view selection. In SurfaceNet+, the rejected occluded views (shown in red) are detected by the projection rays combined with the output point cloud in the previous stage mentioned in subsection\ref{occlusion-aware}.
It is worth noting that these occluded views are extremely hard to detect using only image patches. These patches are  similar to each other, so it is difficult to infer the relative position relationship among them in the absence of three-dimension prior.

\words{In Fig.~\ref{fig:ablation_picture}(b), to further demonstrate the benefit from an explicit ``relative weight (with occlusion-aware)'' ($\bigstar$ curve), we investigate the setting ``relative weight (without occlusion-aware)'' ($\blacktriangle$ curve) and the setting ``without relative weight'' ($\square$ curve). Enabling the ``relative weight (with occlusion-aware)'' significantly improves Recall (Completeness) of the reconstructed model, indicating the effectiveness of the proposed trainable occlusion-aware view selection scheme.

Additionally, in Fig.~\ref{fig:ablation_picture}(a), we evaluate the proposed end-to-end trainable occlusion-aware view selection scheme (trainable, $\bigstar$ curve) versus the heuristic view selection method (heuristic, $\blacksquare$ curve). Note that both of them share the same backbone network structure and the multiscale fusion strategy, while the only difference is the view selection module. As we can see, the proposed end-to-end trainable occlusion-aware view selection scheme significantly boosts the completeness (Recall) of the reconstruction model.
}

\textbf{SurfaceNet\cite{ji2017surfacenet}.} \words{
For fair comparison, in Fig.~\ref{fig:ablation_picture}(a) we also show the performance difference between the ICCV SurfaceNet\cite{ji2017surfacenet} and the modified SurfaceNet with the new backbone, where the only modification on the ICCV SurfaceNet is the network structure that SurfaceNet+ is using (Fig.~\ref{fig:network}). It is worth noting the relative position changes of each curve. There is a clear shift downward after adding the proposed trainable occlusion-aware view selection\textbf{(View-selection)}. This indicates the better recall with comparable precision. Overall, the gain achieved by SurfaceNet+ over SurfaceNet has NO relationship with the backbone network adopted; instead, it is benefited from the proposed multi-scale pipeline and novel view selection strategy.}

\subsection{Discussion}\label{Quantitative analyze}


\textbf{Hyperparameters.} The number of view pairs $N_v$ is also critical for the algorithm. Too few view pairs may lead to noisy and inaccurate reconstruction, while too many in sparse-MVS lead to incomplete (low recall) prediction. The trade-off of $N_v=3$ achieves the best overall performance as indicated in the figure.

\begin{figure}[htbp] 
\centering
\includegraphics[width=0.75\linewidth]{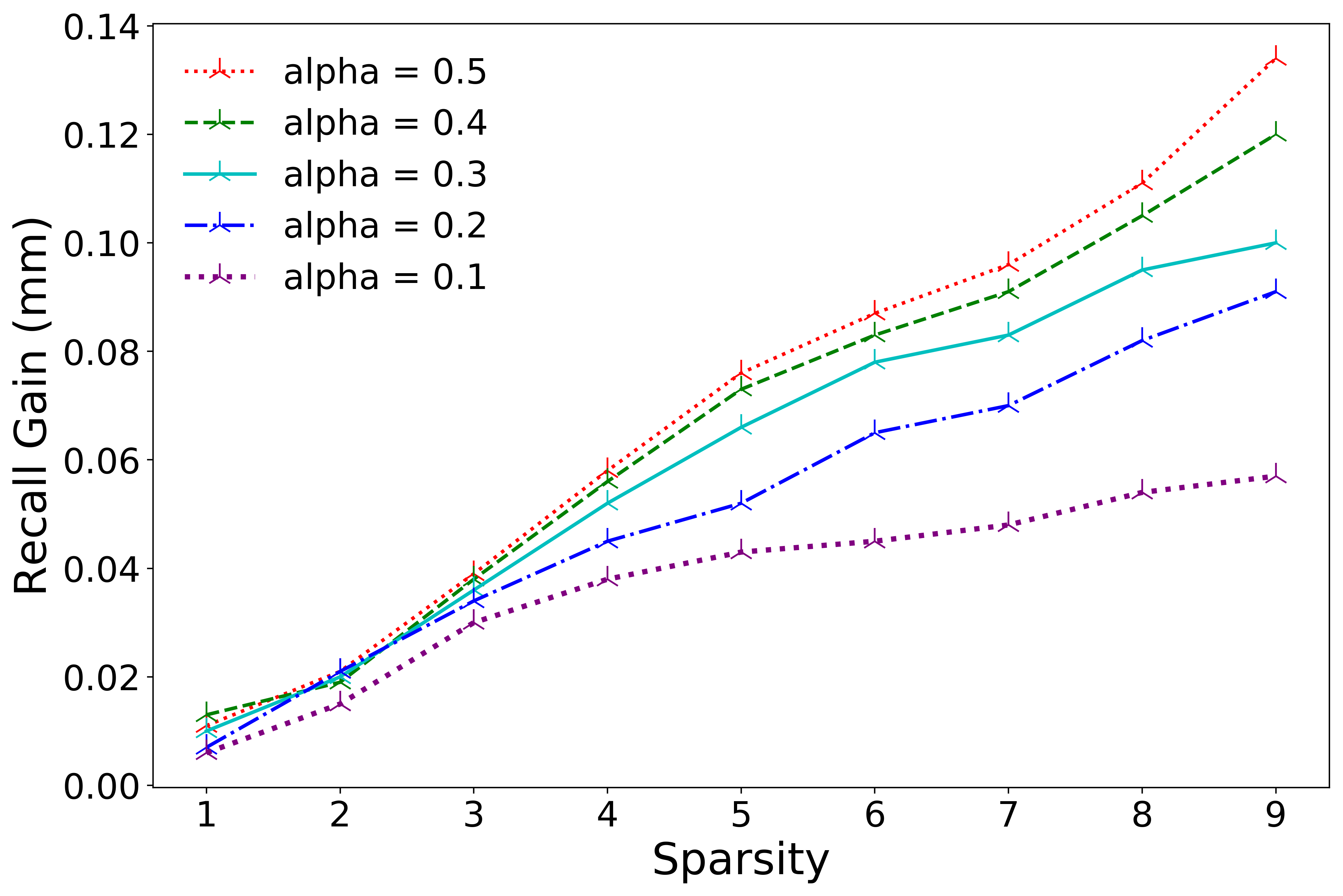} 
\caption{Recall gain w.r.t. $\alpha$(occlusion parameter) at different sparsity. The gain is counted by the recall improvement based on the method without occlusion detection. }
\label{fig:alpha_change}
\end{figure}
To further analyze the effect of occlusion-aware view selection, we experiment with different occlusion parameters $\alpha$ at different sparsities with a fixed view pair number $N_v=3$. The recall gain is counted by the recall improvement based on the method without occlusion detection. 


As shown in Fig.~\ref{fig:alpha_change}, the gain increases as the sparsity grows. The reason lies in that when sparsity increases, a growing baseline angle and fewer view pairs lead to a lower percentage of non-occluded views. Therefore, lower weight on occluded views controlled by alpha has increased benefit on larger sparsity. 


\begin{table}[htbp]
\centering
\small

\begin{tabular}{c|cc|ccc}
    \hline
          & \multicolumn{2}{|c|}{NO. of sub-volumes} & \multicolumn{3}{c}{Speed up} \\ \cline{2-6} 
          & SurfaceNet         & SurfaceNet+        & mean           & max   & min  \\ \hline
    DTU   & 140,608            & 12,320             & \textbf{11X}   & 23X   & 7X   \\
    T\&T & 158,992            & 15,892             & \textbf{15X}   & 33X   & 11X  \\ \hline
\end{tabular}

\captionof{table}{
Efficiency comparison of proposed method with and without volume selection. Where we set the resolution as 4(mm) in the Tanks and Temple dataset (T\&T) and 0.03(mm) in the DTU dataset and compare the average cubic number and its mean and maximum speed up ratio for each model.
} \label{tab:cub_num}
\end{table}

\vspace{3mm}
\begin{figure}[htbp] 
\centering
\includegraphics[width=0.75\linewidth]{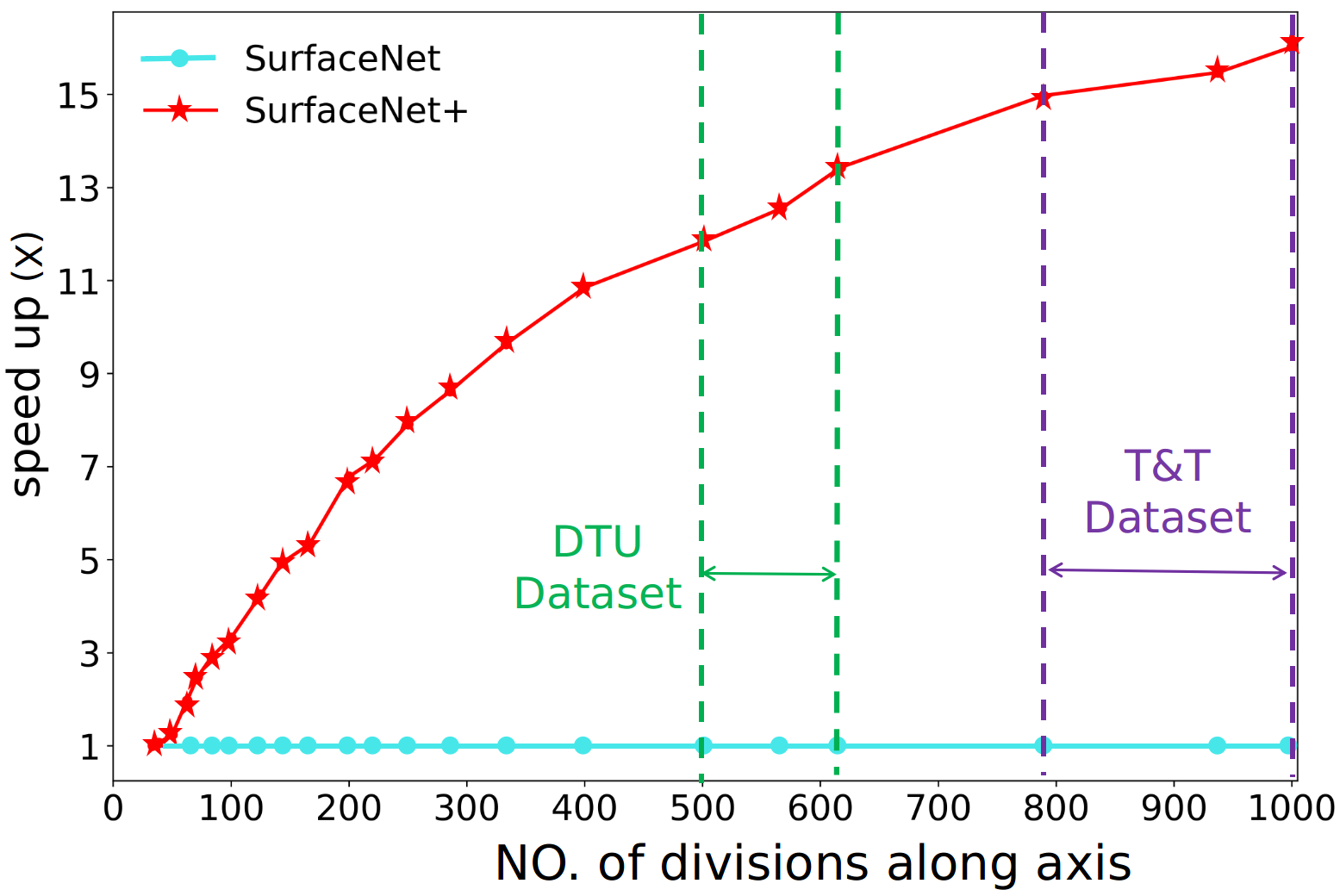} 
\caption{Speed up ratio with the change of resolution. 
Note how coarse to fine mechanism leads to efficient representation compared to SurfaceNet\cite{ji2017surfacenet}. With the finer of reconstruction, speed up ratio grows dramatically. }
\label{fig:cubic_change}
\end{figure}
\noindent\textbf{Efficiency.}
To evaluate the efficiency brought by the coarse-to-fine mechanism, we measure the speed of the algorithm using the total sampled volumes. Specifically, we count each number of cubes sampled by the algorithm for both methods in all the models on the DTU \cite{aanaes2016large} evaluation set and Tanks and Temples \cite{Knapitsch2017} `Intermediate' set. We set the whole reconstruction scene as a cubic box with length $l_{scene} = 400(mm)$ in the DTU dataset and the final voxel resolution $r =  0.3(mm)$. Each volume forms a tensor of size $s \times s \times s$ and we set $s = 64$. We use all the cubes for reconstruction in SurfaceNet \cite{ji2017surfacenet}, a three-stage coarse-to-fine pipeline for SurfaceNet+. The settings in Tanks and Temples are equal to DTU except that we set $l_{scene} = 400(mm)$ and $r =  2(mm)$. The left part of Table~\ref{tab:cub_num} shows the average sub-volumes used for reconstruction, and the right part shows the speed up multiple brought by the coarse-to-fine mechanism. We value the average multiple as the ratio between two methods. The mean, maximum and minimum multiple show that the volume selection mechanism can achieve more than 10 times higher efficiency on both datasets. 


To better understand the efficiency promotion, in Fig.~\ref{fig:cubic_change}, we visualize the speed up ratio as the scale of the relative resolution $r_{relative} = \dfrac{l_{scene}}{r}$ .
Note how the coarse-to-fine mechanism leads to efficient representation compared to SurfaceNet. At low relative resolution, the ratio is near 1
due to the nearly dense sampling based on the coarse prediction. Yet with the finer reconstruction, the speedup ratio grows dramatically because the finer prediction leads to a higher percentage of empty sub-volumes.  




\begin{figure}[htbp] 
\centering
\includegraphics[width=0.7\linewidth]{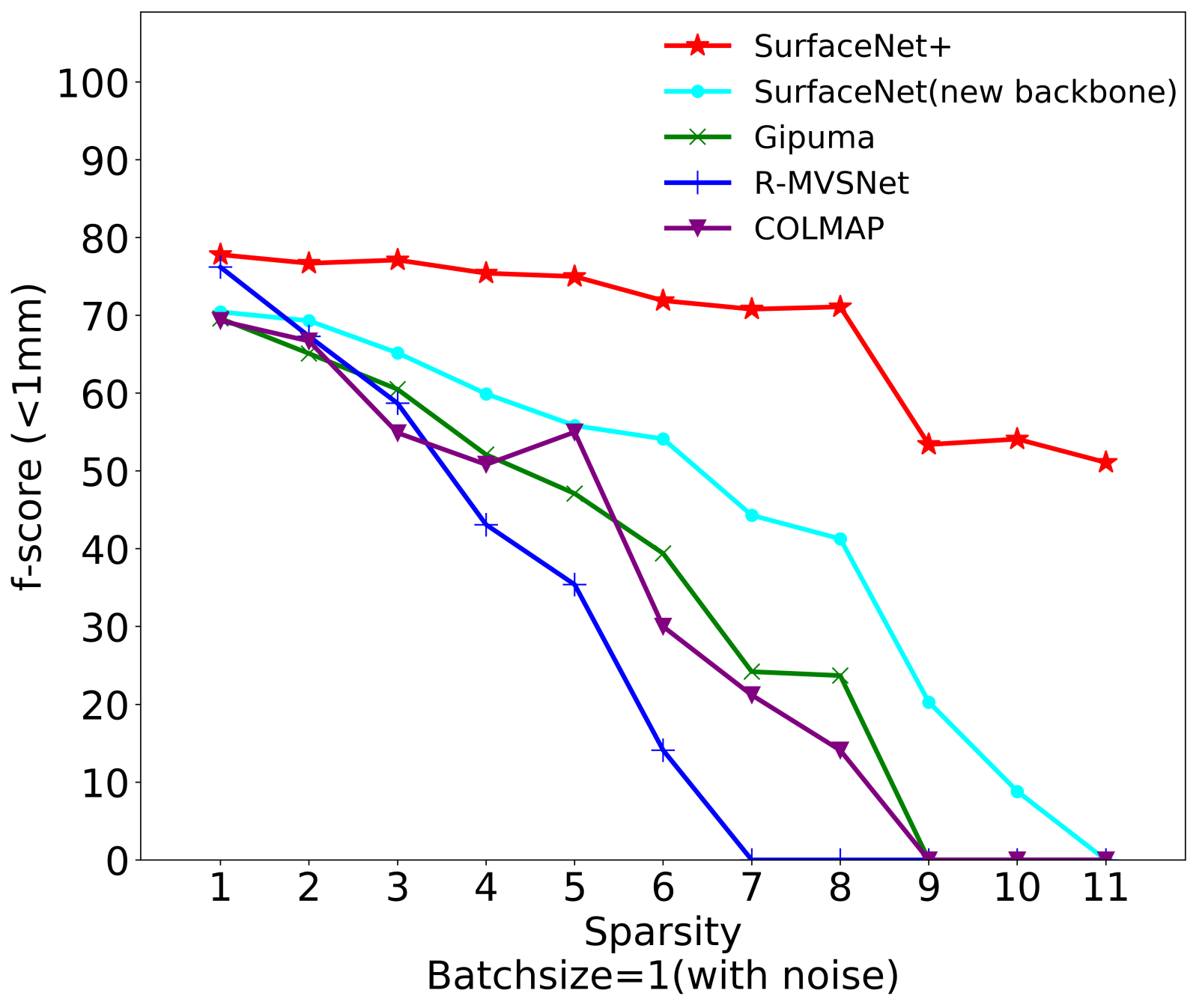} 
\caption{ 
\words{Evaluation of the sparse MVS benchmark using the \textbf{noisy camera poses} (estimated by SfM\cite{schoenberger2016sfm}),}
}
\label{fig:noise_dtu}
\end{figure}

\vspace{3mm}
\noindent\textbf{Noisy camera poses.}
\words{
The camera poses used in our previous experiments are given by the public datasets, which are estimated by the registration of laser scans (denoted as GT camera pose). While in practice, the camera poses may be computed through the sparse set of views, which inevitably suffers noise (denoted as noisy camera pose). 
To evaluate how the noisy camera pose affects the performance of SurfaceNet+, we adopt the structure-of-motion SfM\cite{schoenberger2016sfm} along with the sparse set of views to obtain the noisy camera pose. 
As expected, using Noisy camera poses (Fig.~\ref{fig:noise_dtu}) degrades the performance of MVS methods that using GT camera poses (Fig.~\ref{fig:batch_result}), where the f-score drops. 

We examine the f-score degradation between Fig.~\ref{fig:noise_dtu} and Fig.~\ref{fig:batch_result}, where the image-wise view selection scheme, used in Gipuma\cite{galliani2015massively} and R-MVSNet\cite{yao2019recurrent}, is more sensitive to the camera pose noise, especially under massive sparsity levels. In contrast, the pixel-wise (COLMAP\cite{schoenberger2016mvs}) and volume-wise (SurfaceNet\cite{ji2017surfacenet} and SurfaceNet+) view selection strategy is relatively more robust to camera pose noise. The reason is that the camera pose noise will introduce an inhomogeneous shift of the photo-consistent matches, so that the pixel-wise and volume-wise view selection can adaptively choose the relatively better views based on the photometric consistency despite the noisy camera pose. In contrast, the image-wise view selection leads to matching the correspondence only on the pre-selected views, which no longer be the best views for a large proportion of pixels or sub-volumes if the camera pose noise is considered. 
}


\section{Conclusion}
As \sentences{sparser sensation is more practical and more cost-efficient}, instead of only focusing on dense MVS setup, we propose a comprehensive analysis on sparse-MVS under various observation sparsities. The proposed leader-board calls for more attention and effort from the community to the sparse-MVS problem, since the state-of-the-art depth-fusion methods significantly perform worse as the baseline angle get larger in the sparser setting. As another line of \words{the} solution, we propose a volumetric method, SurfaceNet+, to handle sparse-MVS by introducing the novel occlusion-aware view selection scheme as well as the multi-scale strategy. Consequently, the experiments demonstrate the tremendous performance gap between SurfaceNet+ and the recent methods in terms of precision and recall. Under the extreme sparse-MVS settings in two datasets, where existing methods can only return very few points, SurfaceNet+ still works as well as in the dense MVS setting.

\textbf{Limitations.} (1) Ideally, for \words{a} simple geometric region, each piece of \words{a} surface in sub-volume should be effectively reconstructed \words{only} using ONE view pair with large baseline angle, \ie $N_v=1$. However, due to various of shading and lighting conditions, the colorization of the 3D model gets more challenging by using less number of views. (2) Furthermore, even though the scanned models in the MVS datasets are large-scale scene, it will be challenging for SurfaceNet+ to effectively and efficiently reconstruct a city-level 3D model. (3) Last but not least, despite of the great generalization ability of the learnt model, it still requires dozens of laser-scanned 3D model for supervision. That significantly limits the application scenarios, such as astro-observation and multi-view microscopic observation, where rare supervision signal can be captured.

\IEEEappendices  






\ifCLASSOPTIONcaptionsoff
  \newpage
\fi



\bibliographystyle{IEEEtran}
\bibliography{egbib}
%



%

\newpage
\begin{IEEEbiography}[{\includegraphics[width=1in,height=1.25in,clip,keepaspectratio]{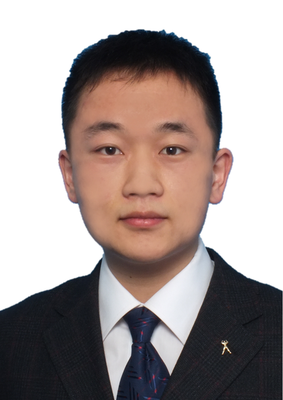}}]{Mengqi Ji} is currently a postdoc in Tsinghua University. He received Ph.D / M.Sc from the Hong Kong University of Science and Technology in 2019 / 2013, and B.E. from University of Science and Technology Beijing in 2012. His research interests include 3D vision and computational photography.
\end{IEEEbiography}
\begin{IEEEbiography}[{\includegraphics[width=1in,height=1.25in,clip,keepaspectratio]{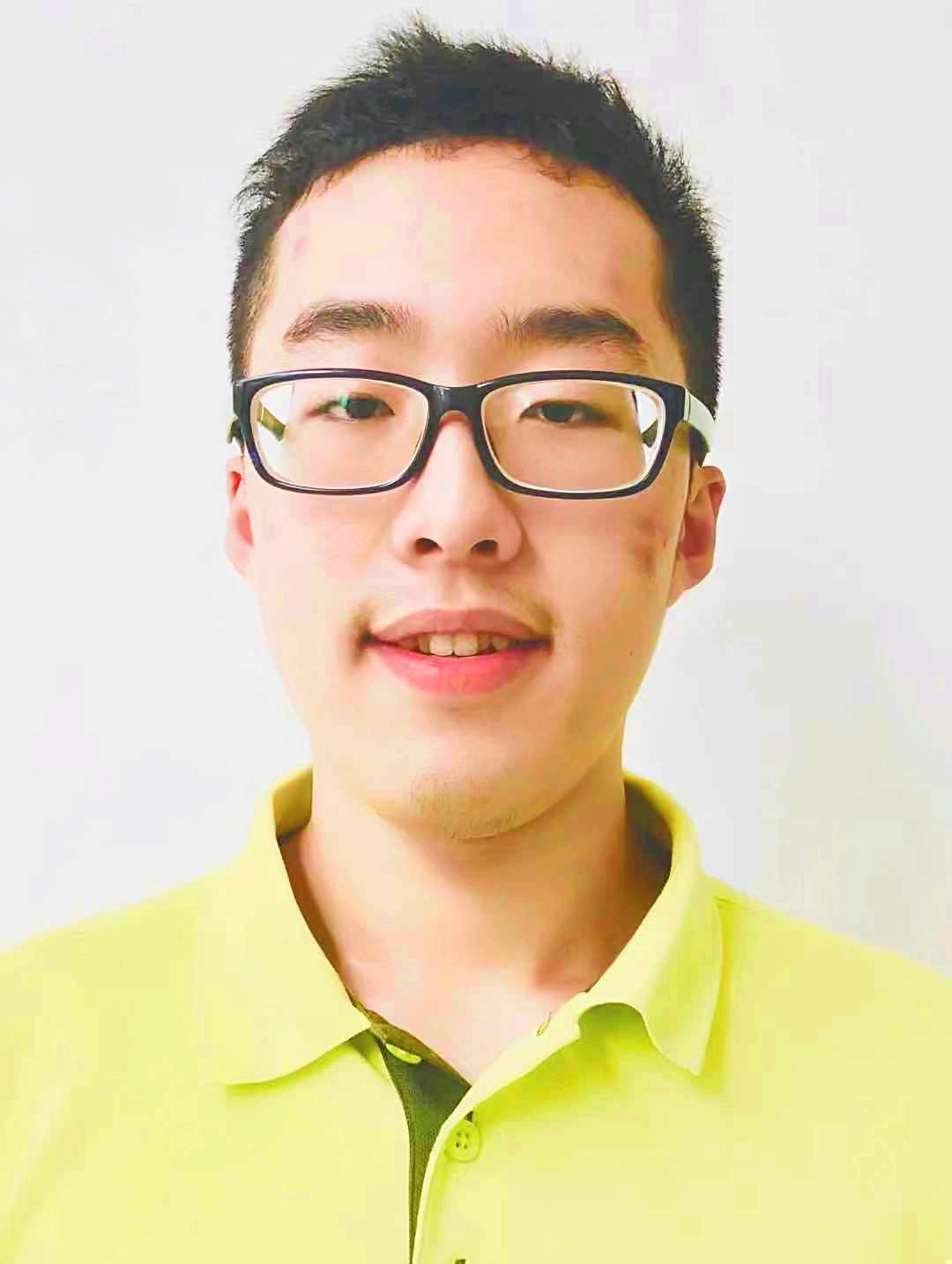}}]{Jinzhi Zhang} is currently a master student in Tsinghua-Berkeley Shenzhen Institute (TBSI), Tsinghua University. He received B.E. from Huazhong University of Science and Technology in 2019. His research interest is 3D vision.
\end{IEEEbiography}


\begin{IEEEbiography}[{\includegraphics[width=1in,height=1.25in,clip,keepaspectratio]{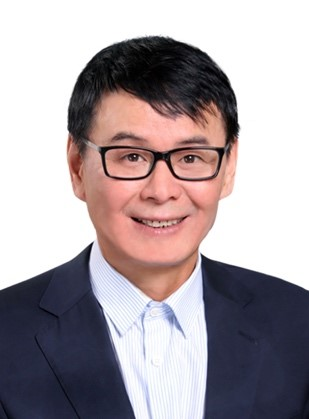}}]{Qionghai Dai} is a professor in the Department of Automation, and an adjunct professor in the School of Life Science, Tsinghua University. Dr. Dai is the academician of Chinese Academy of Engineering. His research interests include computational photography, brain science, and artificial intelligence. Dr. Dai is currently IEEE Senior Member, serving as Associate Editor for IEEE TIP. 
\end{IEEEbiography}

\begin{IEEEbiography}[{\includegraphics[width=1in,height=1.25in,clip,keepaspectratio]{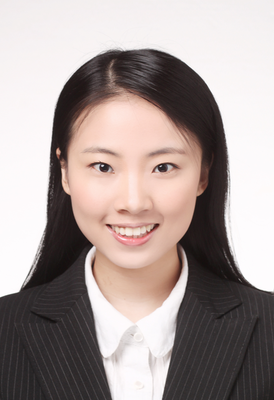}}]{Lu Fang} is currently an Associate Professor in Tsinghua University. She received Ph.D from the Hong Kong Univ. of Science and Technology in 2011, and B.E. from Univ. of Science and Technology of China in 2007. Her research interests include computational photography and 3D vision. Dr. Fang used to receive Best Student Paper Award in ICME 2017, Finalist of Best Paper Award in ICME 2017 and ICME 2011. Dr. Fang is currently IEEE Senior Member.
\end{IEEEbiography}




\end{document}